\pdfoutput=1 
\documentclass[runningheads]{llncs}
\usepackage{graphicx}

\usepackage{tikz}
\usepackage{comment}
\usepackage{amsmath,amssymb} 
\usepackage{color}

\usepackage{subfig}
\usepackage{multirow}
\usepackage{bbm}
\usepackage{ulem}
\usepackage{cite}
\usepackage{mathtools} \usepackage{xspace}           \newcommand{\mbf}{\mathbf{f}}     
\newcommand{\mbn}{\mathbf{n}}
 \newcommand{\mbp}{\mathbf{p}}   \newcommand{\mbt}{\mathbf{t}}  \newcommand{\mbv}{\mathbf{v}}        \newcommand{\mbC}{\mathbf{C}}   \newcommand{\mbF}{\mathbf{F}}   \newcommand{\mbI}{\mathbf{I}}    \newcommand{\mbM}{\mathbf{M}}   \newcommand{\mbP}{\mathbf{P}}  \newcommand{\mbR}{\mathbf{R}}  \newcommand{\mbT}{\mathbf{T}}    \newcommand{\mbX}{\mathbf{X}}                                                                                     

\usepackage[accsupp]{axessibility}  

\newcommand\blfootnote[1]{%
  \begingroup
  \renewcommand\thefootnote{}\footnote{#1}%
  \addtocounter{footnote}{-1}%
  \endgroup
}

\begin{document}
\pagestyle{headings}
\mainmatter
\def\ECCVSubNumber{2418}  

\title{Multi-Person 3D Pose and Shape Estimation \\via Inverse Kinematics and Refinement} 

\titlerunning{Multi-Person 3D Pose and Shape Est. via IK and Refinement}
%
\author{Junuk Cha\inst{1}\orcidID{0000-0003-2321-2797} \and
Muhammad Saqlain\inst{1,2\dag}\orcidID{0000-0001-5877-6432} \and
GeonU Kim\inst{1\ddag} \and
Mingyu Shin\inst{1,3\ddag} \and
Seungryul Baek\inst{1}\orcidID{0000-0002-0856-6880}}
\authorrunning{Cha et al.}
%
\institute{1 UNIST, South Korea
\qquad 2 eSmart Systems, Norway \qquad 3 Yeongnam Univ., South Korea} 
\maketitle

\blfootnote{This research was conducted when Dr. Saqlain was the post-doctoral researcher at UNIST$\dag$, and when Mr. Kim and Mr. Shin were undergraduate interns at UNIST$\ddag$.}

\begin{abstract}
Estimating 3D poses and shapes in the form of meshes from monocular RGB images is challenging. Obviously, it is more difficult than estimating 3D poses only in the form of skeletons or heatmaps. When interacting persons are involved, the 3D mesh reconstruction becomes more challenging due to the ambiguity introduced by person-to-person occlusions. To tackle the challenges, we propose a coarse-to-fine pipeline that benefits from 1) inverse kinematics from the occlusion-robust 3D skeleton estimation and 2) Transformer-based relation-aware refinement techniques. In our pipeline, we first obtain occlusion-robust 3D skeletons for multiple persons from an RGB image. Then, we apply inverse kinematics to convert the estimated skeletons to deformable 3D mesh parameters. Finally, we apply the Transformer-based mesh refinement that refines the obtained mesh parameters considering intra- and inter-person relations of 3D meshes. Via extensive experiments, we demonstrate the effectiveness of our method, outperforming state-of-the-arts on 3DPW, MuPoTS and AGORA datasets.

\keywords{Multi-person, 3D mesh reconstruction, Transformer}
\end{abstract}

\section{Introduction}

Recovering 3D human body meshes for a single person or multi-person from a monocular RGB image has made great progress in recent years~\cite{bogo2016keep,choi2021beyond,choi20213Dcrowdnet,guler2019holopose,jiang2020coherent,kanazawa2018end,kanazawa2019learning,kocabas2020vibe,kocabas2021pare,kocabas2021spec,kolotouros2019learning,lin2021end,lin2021mesh,sun2021monocular,tran2022meshletemp,xu2019denserac,zhang2021body,zhang2021pymaf}. The technique is essential to understand people's behaviors, intentions and person-to-person interactions. It has a wide range of real-world applications such as human motion imitation~\cite{liu2019liquid}, virtual try on~\cite{mir2020learning}, motion capture~\cite{mehta2020xnect}, action recognition~\cite{cha2022learning,saqlain20223dmesh,varol2021synthetic}, etc.

Recently, deep convolutional neural network-based mesh reconstruction methods~\cite{cha2021towards,choi2021beyond,choi20213Dcrowdnet,guler2019holopose,jiang2020coherent,kanazawa2018end,kanazawa2019learning,kocabas2020vibe,kocabas2021pare,kocabas2021spec,kolotouros2019learning,lin2021end,lin2021mesh,sun2021monocular,tran2022meshletemp,xu2019denserac,zhang2021body,zhang2021pymaf} have shown the practical performance on in-the-wild scenes~\cite{ionescu2013human3,joo2015panoptic,von2018recovering,mehta2017monocular}. Most of the existing 3D human body pose and shape estimation approaches~\cite{cha2021towards,choi2021beyond,guler2019holopose,kanazawa2018end,kanazawa2019learning,kocabas2020vibe,kocabas2021pare,kocabas2021spec,kolotouros2019learning,lin2021end,lin2021mesh,xu2019denserac} achieved promising results for single-person cases. Generally, firstly they crop the area with a person in an input image using bounding-box and then extract features for each detected person, which are further used for 3D human mesh regression.

\begin{figure*}[t]
\captionsetup[subfigure]{labelformat=empty}
\centering
\begin{minipage}[b]{.78\textwidth}
\subfloat[{(a)}]{\includegraphics[width=0.24\textwidth]{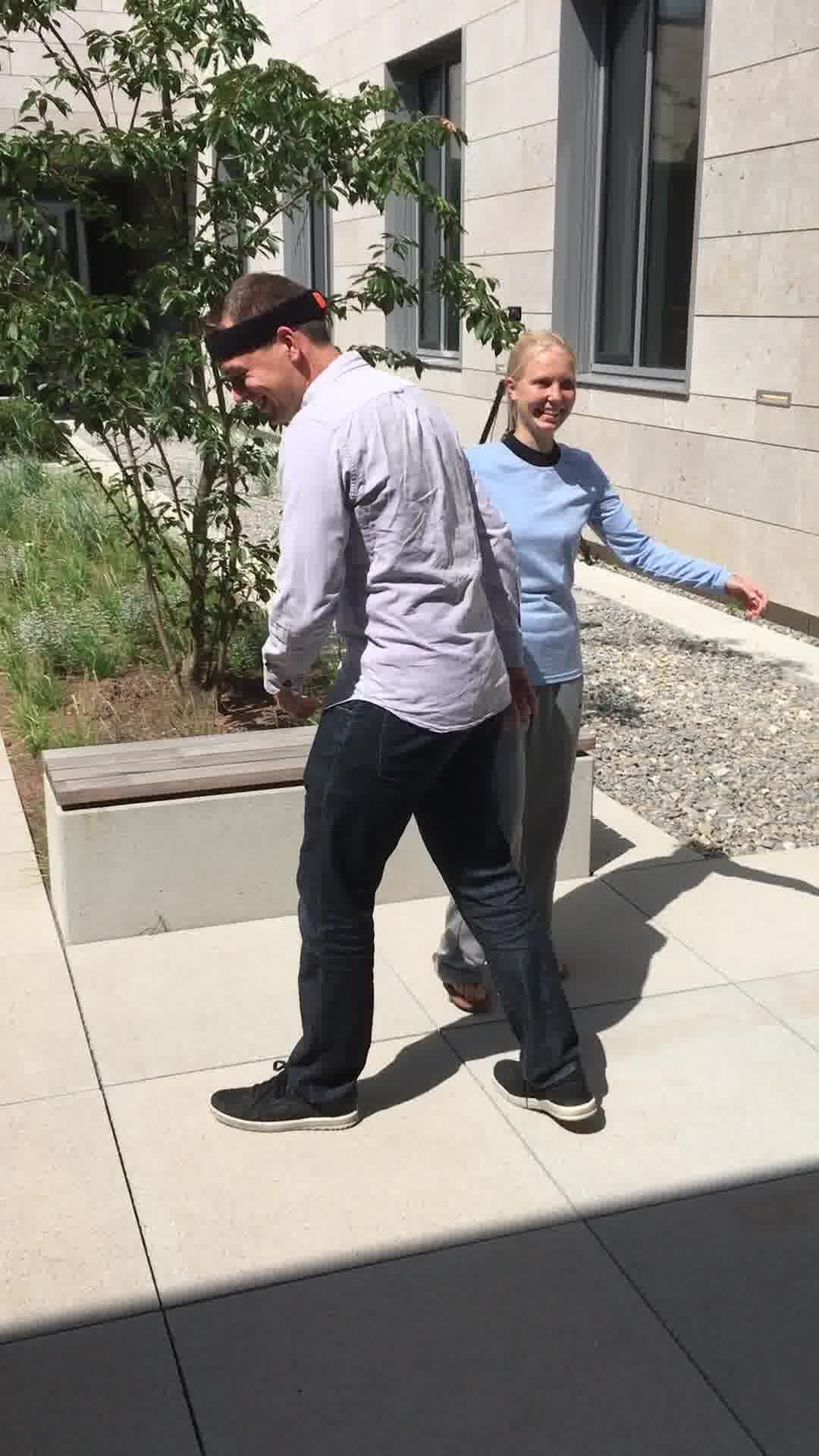}}
\subfloat[][{(b)}]{\includegraphics[width=0.24\textwidth]{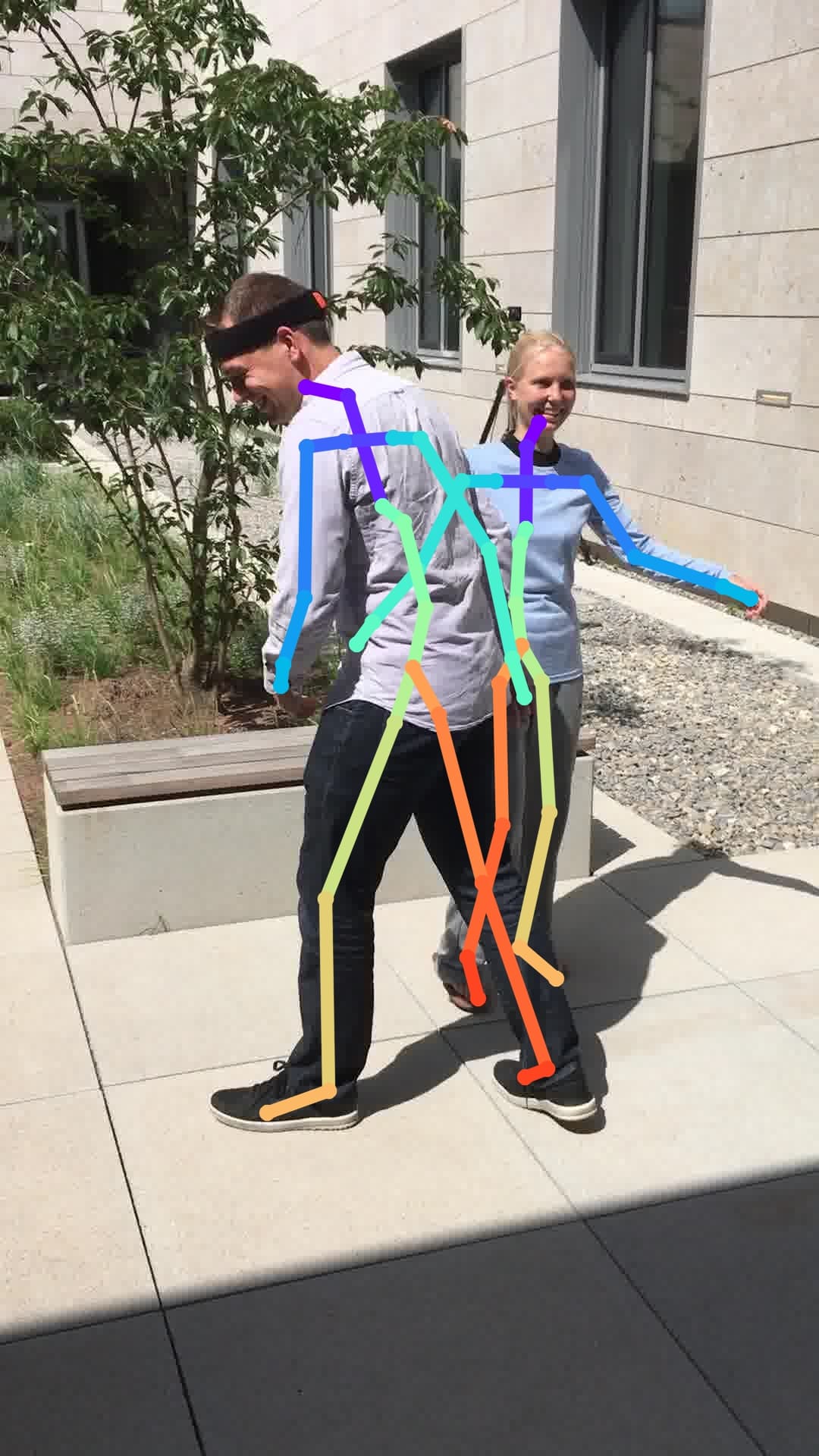}}
\subfloat[][{(c)}]{\includegraphics[width=0.24\textwidth]{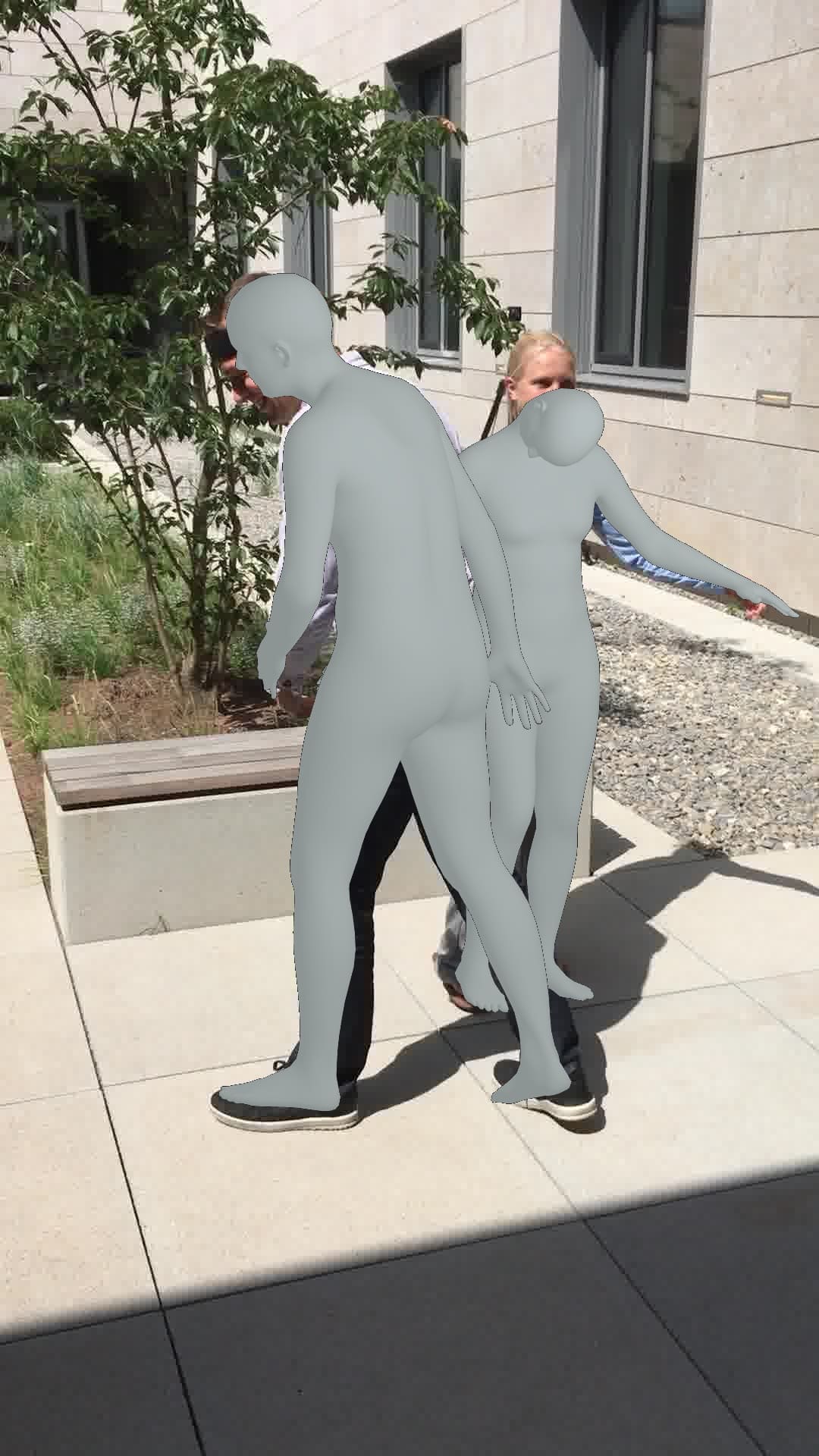}}
\subfloat[][{(d)}]{\includegraphics[width=0.24\textwidth]{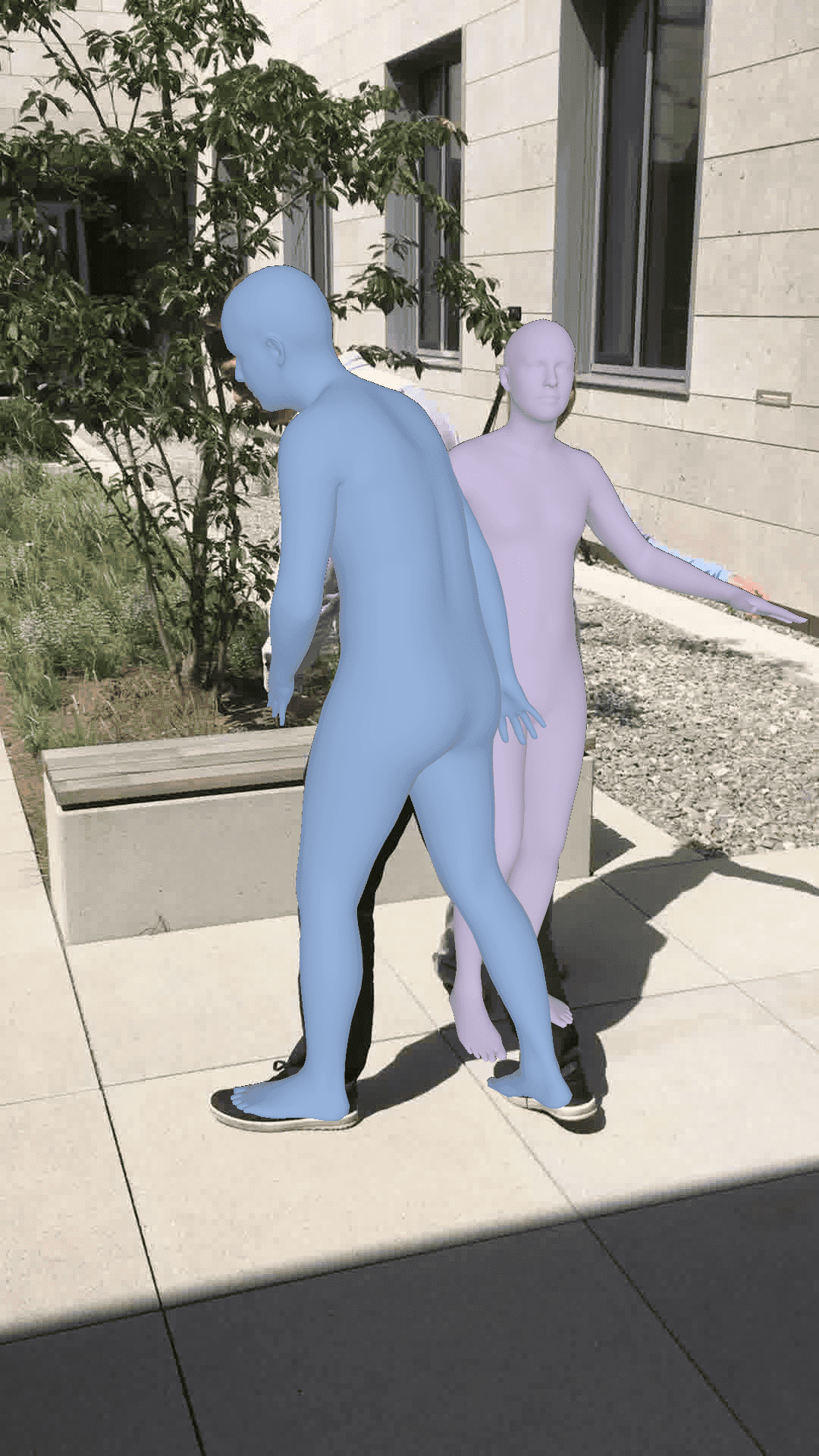}}
\end{minipage}
\begin{minipage}[b]{.18\textwidth}\centering
\subfloat[][{(e)}]{\includegraphics[width=0.98\textwidth,height=0.7\textwidth]{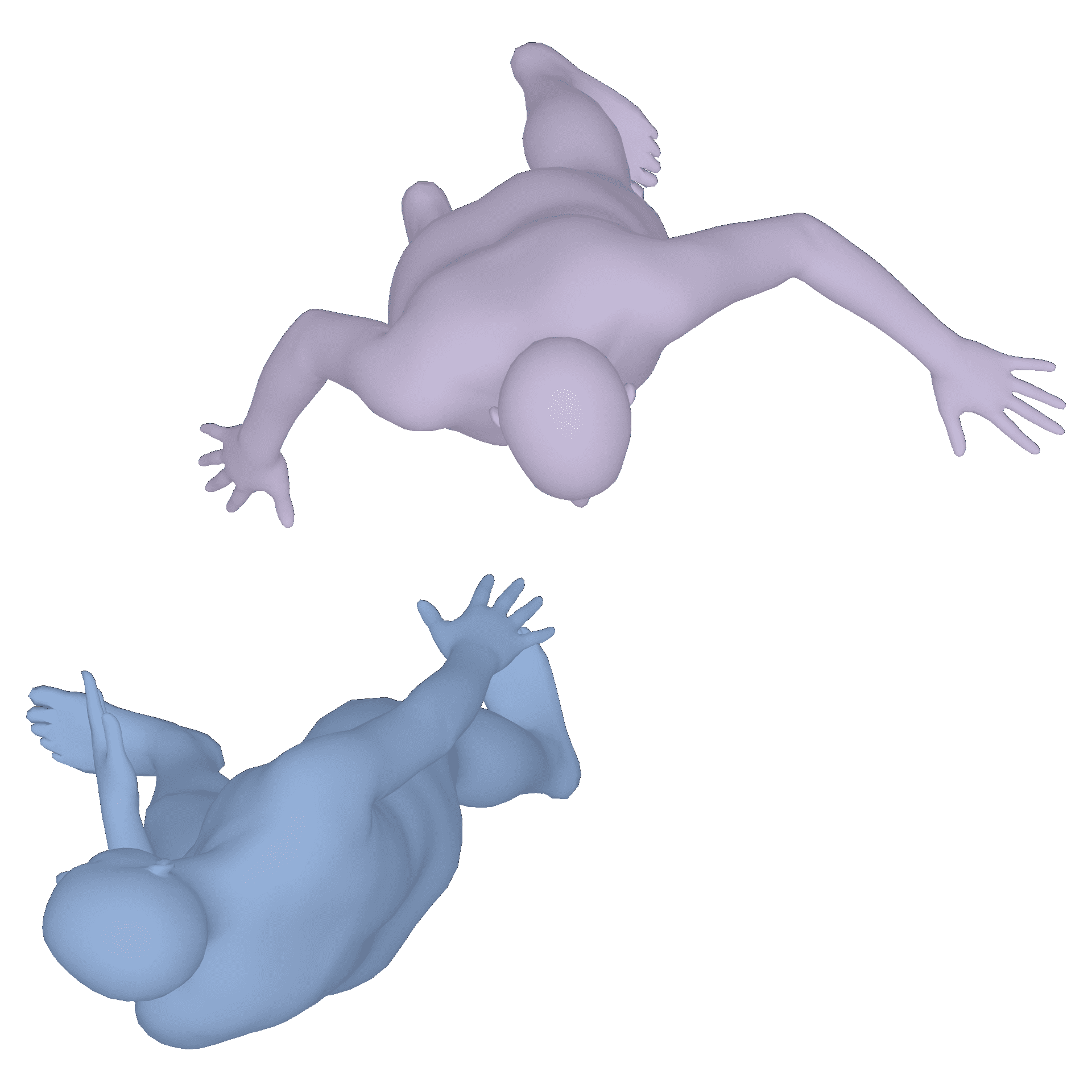}}\\
\subfloat[][{(f)}]{\includegraphics[width=0.98\textwidth,height=0.7\textwidth]{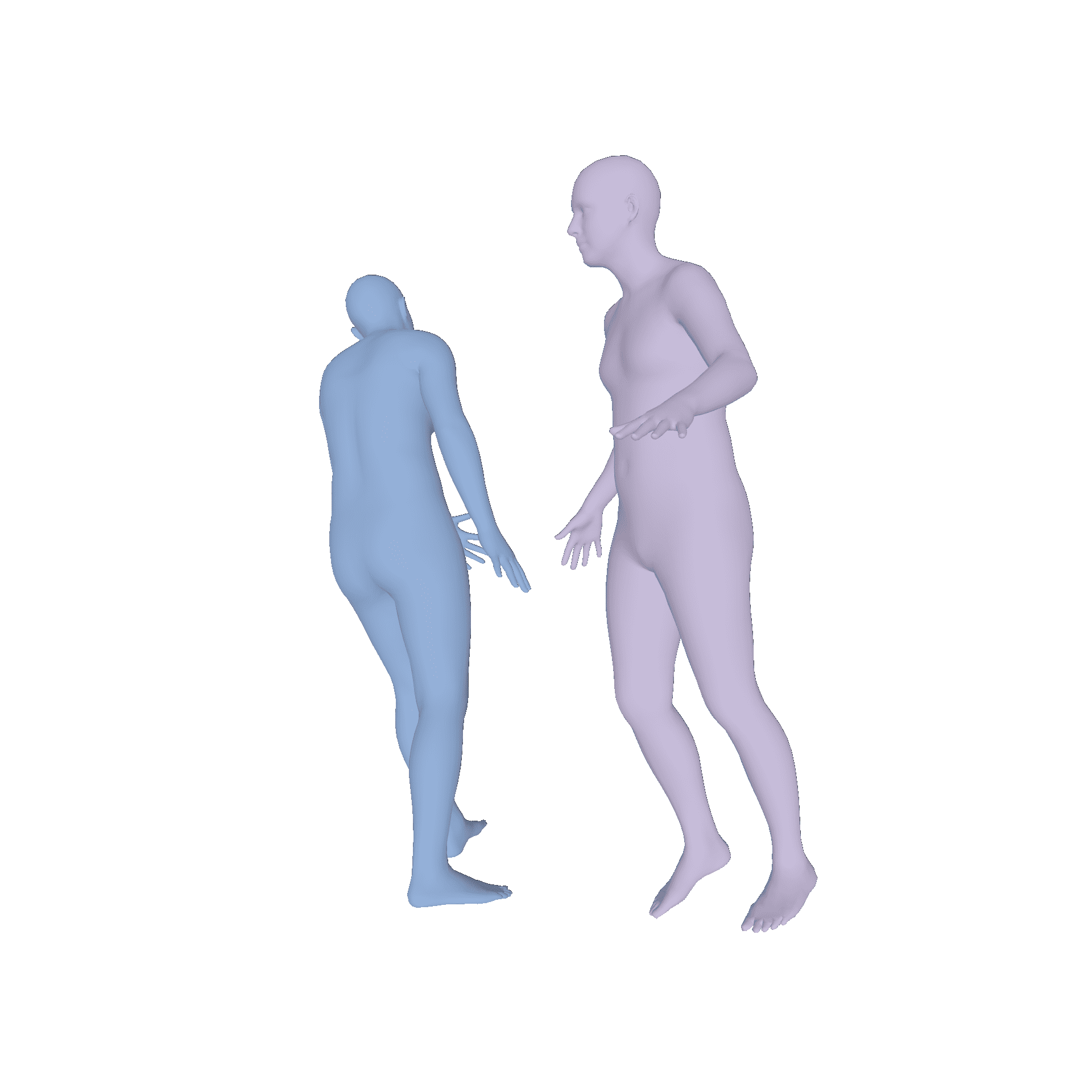}}
\end{minipage}
\caption{Example outputs from our pipeline: (a) input RGB image, (b) initial skeleton estimation results obtained from the input image, (c) initial meshes obtained from the inverse kinematics process, (d) refined meshes obtained from the refinement Transformer, (e, f) top- and side-views for the refined meshes.}
\label{fig1}
\end{figure*}

Some of the recent studies \cite{joo2021exemplar,kanazawa2018end,kanazawa2019learning,kocabas2020vibe,kocabas2021pare,kocabas2021spec,kolotouros2019learning,li2021hybrik,lin2021mesh,tran2022meshletemp,zhang2021pymaf} reconstruct each person 3D mesh individually for multi-person 3D mesh reconstruction using the same bounding-boxes detector~\cite{openpose,he2017mask,redmon2016you}. Multiple persons can create severe person-to-person or person-to-environmental occlusions, erroneous monocular depth and diverse human body appearance which results in performance ambiguity in crowded scenes, while in these methods, proper modules that tackle the interacting persons have not been established yet. A few recent methods~\cite{jiang2020coherent,zhang2021body} applied direct regression for multiple persons which do not require individual person detection. Sun et al.~\cite{sun2021monocular} used body center heatmaps as the target representation to identify mesh parameter map. However, without applying the human detection, the human pose estimation result is frequently affected by unimportant pixels and it frequently fails to capture scale variations, which result in the inferior performance.

In parallel, there have been efforts to reduce the ambiguity of estimating 3D meshes from an RGB image. However in the aspect of the pose recovery, 3D body mesh recovery methods~\cite{kanazawa2018end,kocabas2020vibe,kocabas2021pare,kolotouros2019learning} still fall behind the 3D skeleton or heatmap estimation methods~\cite{cheng20203d,cheng2019occlusion,iskakov2019learnable,sun2018integral}. One drawback of 3D skeleton estimation method is that it cannot reconstruct the full 3D body mesh. Recently, Li et al.~\cite{li2021hybrik} proposed an inverse kinematics method for single-person mesh reconstruction to recover 3D meshes from 3D skeletons. This approach is promising since it is able to deliver good poses obtained from 3D skeleton estimator to the 3D mesh reconstruction pipeline.

To tackle the multi-person 3D body mesh reconstruction task, we propose a coarse-to-fine pipeline that first estimates 3D skeletons, reconstruct 3D meshes from 3D skeletons via inverse kinematics and refine the initial 3D mesh parameters via relation-aware refinement. Inspired by~\cite{sarandi2020metrabs}, our 3D skeleton estimator involves metric-scale heatmaps and is trained by both relative and absolute positional 3D poses to be robust to occlusions. By extending the IK process~\cite{li2021hybrik} towards the multi-person scenario, we are able to obtain the initial 3D meshes for multiple persons from 3D skeletons; while the accuracy is limited especially for interacting person cases. To compensate for the limitation, we propose the relation-aware Transformer to refine the initial mesh parameters considering intra- and inter-person 3D mesh relationships. The Fig.~\ref{fig1} shows example outputs for intermediate steps. To summarize, our contributions are as follows:
\begin{itemize}
\item We propose a coarse-to-fine multi-person 3D body mesh reconstruction pipeline that first estimates 3D skeletons and then delivers it toward 3D meshes via inverse kinematics. To make our pipeline robust to interacting persons, we borrowed the occlusion-robust techniques for 3D skeleton estimation.
\item To further boost the performance, we propose the Transformer-based architecture for relation-aware mesh refinement to refine the initial mesh parameters considering intra- and inter-person relationships.
\item Extensive comparisons are conducted involving three challenging multi-person 3D body pose benchmarks (i.e. 3DPW, MuPoTS and AGORA) and we have demonstrated the state-of-the-art performance on each benchmark. Via ablation studies, we prove that each component works in the meaningful way.
\end{itemize}

\section{Related Works}

\noindent \textbf{Single-person 3D mesh regression.} There is a long history of methods for predicting 3D human body meshes from monocular RGB images or video frames~\cite{guan2009estimating}. Recently, there has been quick advancement in this field thanks to SMPL~\cite{smpl} which provides a low dimensional parameterization of the 3D human body mesh. Here we focus on a 3D body mesh regressing by adopting a parametric model like SMPL from a monocular RGB image. Bogo et al.~\cite{bogo2016keep} represented an optimization-based method called SMPLify by fitting SMPL on the detected 2D body joints iteratively. However, this optimization-based approach is comparatively time-consuming and struggle with the higher inference time per input frame.

Some recent studies~\cite{kolotouros2019convolutional,omran2018neural,pavlakos2018learning} use deep neural networks for SMPL parameters regression from images in a two-stage manner, which have been effective and can generate more accurate mesh reconstruction outputs in the presence of large-scale 3D datasets. They first determine intermediate renderings such as silhouettes and 2D keypoints from input images and then map them to the SMPL parameters. Impressive results have been achieved for in-the-wild images by applying diverse weak supervision signals such as semantic segmentation \cite{xu2019denserac}, texture consistency \cite{pavlakos2019texturepose}, efficient temporal features \cite{kocabas2020vibe,sun2019human,tung2017self}, 2D pose \cite{choi2020pose2mesh,kanazawa2018end,kundu2020appearance}, motion dynamics \cite{kanazawa2019learning}, etc.

More recently, Li et al. \cite{li2021hybrik} proposed a 3D human body pose and shape estimation method by collaborating the 3D keypoints and body meshes. Authors introduced an inverse kinematics process to find the relative rotations using twist-and-swing decomposition which estimates targeted body joint locations.

\noindent \textbf{Multi-person 3D skeleton regression.} There have been variety of methods~\cite{dong2021shape,mehta2018single,moon2019camera,rogez2017lcr} that tackle the 3D body pose estimation for multi-person: Zanfir et al.~\cite{rogez2017lcr} proposed LCR-Net that consists of localization, classification, and regression modules. The localization module detects multi-persons from a single image. The classification module classifies the detected human into several anchor-poses. Finally, the regression module refines the anchor-poses. Mehta et al.~\cite{mehta2018single} proposed a single-shot method for multi-person 3D pose estimation from a single image. In addition, they introduced the MuCo-3DHP dataset which has multi-person interactions and occlusions images. Moon et al.~\cite{moon2019camera} proposed top-down method for 3D multi-person pose estimation from a monocular RGB image. This method consists of human detection, absolute 3D human root localization, and root-relative 3D single-person pose estimation modules. Dong et al.~\cite{dong2021shape} used the multi-view images for estimating the multi-person 3D pose. They proposed a coarse-to-fine method lifting the 2D joints to the 3D joints. They obtained the 2D joints candidates from \cite{openpose}. The initial 3D joints are triangulated from 2D joints candidates of different camera views of the same image. In addition, the initial 3D joints are updated using the prior information using the SMPL~\cite{smpl} model.

Recent multi-person 3D pose regression works~\cite{cheng2021monocular,reddy2021tessetrack,sarandi2020metrabs,zhang2021direct} tackled a variety of issues such as developing attention-based mechanism dedicated to the 3D pose estimation problem which considers 3D-to-2D projection process~\cite{zhang2021direct}, combining the top-down and bottom-up networks~\cite{cheng2021monocular}, developing the tracking-based for multi-person~\cite{reddy2021tessetrack} and so on. S{\'a}r{\'a}ndi et al.~\cite{sarandi2020metrabs} recently proposed a metric-scale 3D pose estimation method that is robust to truncations. It is able to reason about the out-of-image joints well. Also, this method is robust to occlusion and bounding-box noise.

\noindent \textbf{Multi-person 3D mesh regression.} There have been few works~{\cite{choi20213Dcrowdnet,jiang2020coherent,sun2021monocular,sun2021putting,zanfir2018deep,zhang2021body}} that concern the multi-person 3D body mesh regression: The approaches could be categorized into two: bottom-up and top-down methods. 

Bottom-up methods~\cite{jiang2020coherent,sun2021monocular,sun2021putting,zhang2021body} perform multi-person detection and 3D mesh reconstruction simultaneously. Zhang et al. \cite{zhang2021body} proposed a Body Mesh as Point (BMP) using a multi-scale 2D center map grid-level representation, which locates selective persons at the  grid cell's center. Sun et al.~\cite{sun2021monocular} proposed a ROMP, which creates parameter maps (i.e. body center heatmap, camera map and mesh parameter map) for 2D human body detection, body positioning and 3D body mesh parameter regression, respectively. Jiang et al.~\cite{jiang2020coherent} represented a coherent reconstruction of multiple humans (CRMH) model, which utilizes the Faster R-CNN based RoI-aligned feature of all persons to estimate SMPL parameters. They further defined the position relevance between multiple persons through a depth ordering-aware loss and an interpenetration. Sun et al.~\cite{sun2021putting} further introduced Bird’s-Eye-View (BEV) representation for reasoning the multi-person body centers and depth simultaneously and combining them to estimate 3D body positions. 

Top-down methods \cite{choi20213Dcrowdnet,zanfir2018deep} first detect each individual person in the frame using bounding-boxes and then estimate the 3D mesh parameters of each detected person. They are basically similar to the single-person 3D mesh reconstruction pipeline; however different in that they provide dedicated modules or loss functions for the multi-person scenario. For example, Zanfir et al.~\cite{zanfir2018deep} proposed a 3D mesh reconstruction method to firstly infer 3D skeletons of each person and group estimated skeletons to infer the final 3D meshes for multi-person. Choi et al.~\cite{choi20213Dcrowdnet} proposed a method for combining early-stage image features and estimated 2D pose heatmaps which are robust to occlusions, to reconstruct 3D meshes for multiple persons.

Bottom-up methods are frequently affected by unimportant image pixels and suffer from scale variations. They further fail to detect small persons since the person detection is not powerful enough compared to that of top-down methods. On the contrary, in top-down methods, proper modules that tackle the interacting persons have not been established yet. In this paper, we take the top-down approach to secure the robustness and propose to use the Transformer architecture to consider the interacting scenario.

\section{Method}

\begin{figure*}[t]
\centering
\includegraphics[width=1\textwidth]{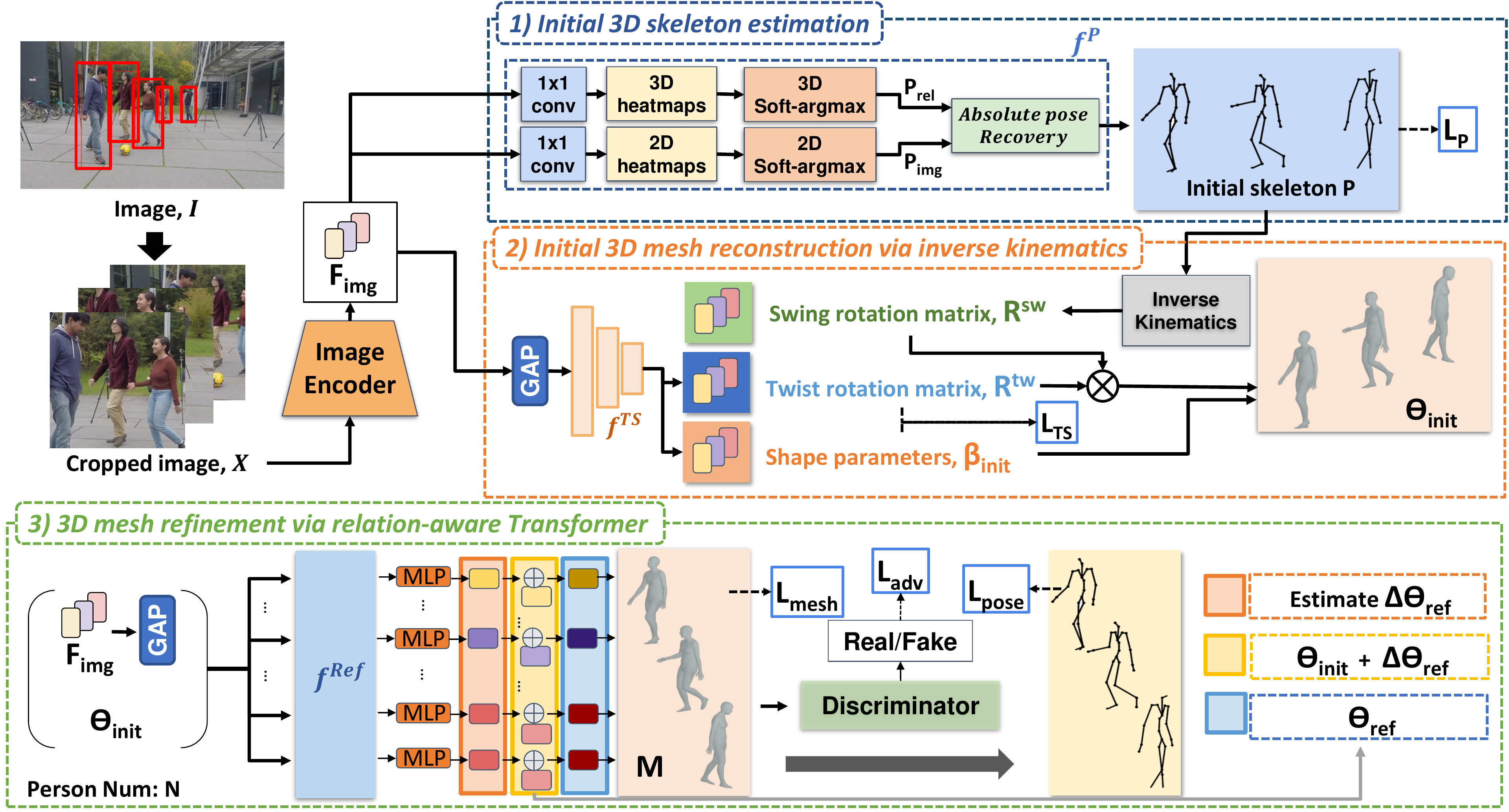}
\caption{The schematic diagram of our framework: We first detect persons from an image $\mbI$ and crop it to $\mbX$ and the image encoder extracts image features $\mbF_\text{img}$ from $\mbX$. Then, initial 3D skeletons $\mbP$ are estimated via the initial 3D skeleton estimator $f^\text{P}$ and SMPL parameters $\boldsymbol{\Theta}_\text{init}$ are reconstructed via the inverse kinematic process, involving the twist angle and shape estimator $f^\text{TS}$ (GAP denotes global average pooling layer). Finally, we refine the initial SMPL parameters by inputting the image features $\mbF_\text{img}$ and $\boldsymbol{\Theta}_\text{init}$ to the relation-aware refiner $f^\text{Ref}$ to produce the refined mesh parameters $\boldsymbol{\Theta}_\text{ref}$. The final 3D mesh $\mbM$ is obtained from the refined SMPL parameters $\boldsymbol{\Theta}_\text{ref}$. The blue boxes denote involved loss functions ($L_\text{P}$, $L_\text{TS}$, $L_\text{mesh}$, $L_\text{adv}$ and $L_\text{pose}$). } 
\label{fig:pipeline}
\end{figure*}

Our aim is to reconstruct 3D meshes $\{\mbM^i\}_{i=1}^M$ of the multiple persons in an RGB image $\mbI$, where $M$ denotes the number of persons in $\mbI$. To achieve this goal, we propose the coarse-to-fine reconstruction pipeline as in Fig.~\ref{fig:pipeline} that 1) first estimates the 3D skeletons $\{\mbP^i\}_{i=1}^M$, 2) obtains the deformable 3D mesh parameters from 3D skeletons via the inverse kinematics (IK) process and 3) refines the initial 3D meshes $\{\mbM^i\}_{i=1}^M$ using the Transformer architecture that considers intra-person and inter-person relationships. In the remainder of this section, we will elaborate each step in detail.

\noindent \textbf{SMPL body model.} For the 3D mesh representation, we use the SMPL deformable 3D mesh model~\cite{smpl} for its compact representation. Variations of the SMPL model~\cite{smpl} are controlled by pose ${\boldsymbol{\theta}}\in\mathbb{R}^{24\times 6}$ and shape parameters ${\boldsymbol{\beta}}\in\mathbb{R}^{1\times10}$. The pose and shape parameters contain 3D rotational information of 24 human body joints in 6D representation and the top-$10$ principal component analysis coefficients of the 3D shape space, respectively. Using the differentiable mapping between SMPL parameters (i.e. ${\boldsymbol{\theta}}$ and ${\boldsymbol{\beta}}$) and the 3D body mesh $\mbM=\{\mbv, \mbf\}$ defined in~\cite{smpl}, we can differentiably obtain the 3D body mesh $\mbM$ from $\boldsymbol{\theta}$ and $\boldsymbol{\beta}$, where $\mbv\in\mathbb{R}^{6,890\times3}$, $\mbf\in\mathbb{R}^{13,776\times 3}$ denote vertices having $6,890$ vertices, $13,776$ triangular faces that are defined by 3 vertices. 

\subsection{Initial 3D Skeleton Estimation}
\label{method:skeleton-estimation}

We take the top-down approach for 3D skeleton estimation that first detect bounding boxes of the humans and estimate 3D skeletons within each bounding box. Following~{\cite{ludl2020enhancing,ning2020lighttrack,sarandi2018synthetic,sarandi2020metrabs}}, we constituted the person detector using the YOLOv4~{\cite{bochkovskiy2020yolov4}} to obtain the cropped image $\mbX\in\mathbb{R}^{256\times 256\times 3}$ from an image $\mbI$. In order to develop the initial 3D skeleton estimation network $f^\text{P}:\mbX\rightarrow \mbP\in\mathbb{R}^{K\times3}$ aiming to use it for the inverse kinematics (IK) process, it is necessary to align the output dimension $K$ with the SMPL model~\cite{smpl}: We are required to set $K$ as $24$ to align it with the SMPL model which uses $24$-dimensional pose parameters. The reason behind this is that the IK process we use (see Sec.~\ref{method:Inverse-Kinematics}) requires to calculate the SMPL pose parameters ${\boldsymbol{\theta}}$ by comparing the 3D skeletons to the SMPL template skeletons having $24$ joints.

To further obtain occlusion-robust 3D skeletons, we follow the architecture and loss functions of the recent 3D skeletal estimation approach~\cite{sarandi2020metrabs} which utilizes the metric-scale heatmaps for 3D skeleton estimation and use both absolute-scale 3D skeletons and image aligned skeletons as the supervision. Within $f^\text{P}$, we applied ResNet~\cite{he2016identity} as a feature extractor that predicts image features $\mbF_\text{img}\in\mathbb{R}^{8\times 8\times 2,048}$. They are fed to a 1x1 convolutional layer to extract 3D heatmaps that can produce the root-relative 3D skeletons $\mbP_\text{rel}$. In parallel, the image features $\mbF_\text{img}$ are fed to a 1x1 convolutional layer to obtain image-scale 2D heatmaps which further produces the 2D skeletons $\mbP_\text{img}$. Finally, the absolute 3D skeletons $\mbP$ are differentiably calculated by combining $\mbP_\text{img}$ and $\mbP_\text{rel}$ with camera intrinsics as in~\cite{sarandi2020metrabs}.

\subsection{Initial 3D Mesh Reconstruction via Inverse Kinematics.} \label{method:Inverse-Kinematics}

We define the inverse kinematics (IK) as the process that reveals angle $\boldsymbol{\theta}$ and shape $\boldsymbol{\beta}$ parameters of the SMPL~\cite{smpl} model from estimated 3D skeletons $\mbP$. The angle parameter $\boldsymbol{\theta}$ could be obtained by finding the relative rotation matrix $\mbR$ that rotates the template skeletons $\mbT=\{\mbt_k\}^K_{k=1}$ to locate it on estimated initial 3D skeletons $\mbP={\{\mbp_k}\}^K_{k=1}$. To reconstruct this, we use the same formula as \cite{li2021hybrik} that decompose the relative rotation matrix with twist and swing angles.

\noindent \textbf{Reconstructing swing angles $\boldsymbol{\alpha}$.} The axis of swing rotation $\mbn_k$ which is perpendicular to $\mbt_k$ and $\mbp_k$ and the swing angle $\boldsymbol{\alpha}=\{\alpha_k\}^K_{k=1}$ are expressed as:
\begin{eqnarray}
    \mbn_k = \frac{\mbt_k\times\mbp_k}{\|\mbt_k\times\mbp_k\|},\quad \cos{\alpha_k}=\frac{\mbt_k\cdot\mbp_k}{\|\mbt_k\|\|\mbp_k\|},\quad \sin{\alpha_k}=\frac{\mbt_k\times\mbp_k}{\|\mbt_k\|\|\mbp_k\|}
    \label{eq:swing}
\end{eqnarray}
By the Rodrigues formula, swing rotation matrix $\mbR^\text{sw}_k$ can be derived from the axis $\mbn_k$ and the angle $\alpha_k$ as follows:
\begin{eqnarray}
    \mbR^\text{sw}_k = \textbf{I} + \sin{\alpha_k}[\mbn_k]_\times + (1-\cos{\alpha_k})[\mbn_k]^2_\times
\end{eqnarray}
where $\textbf{I}$ is $3\times 3$ identity matrix and $[\mbn_k]_\times$ is the skew-symmetric matrix of $\mbn_k$. 

\noindent \textbf{Reconstructing twist angles $\boldsymbol{\phi}$ and shape parameter $\boldsymbol{\beta}$.} While swing angles $\boldsymbol{\alpha}$ could be obtained from $\mbt_k$ and $\mbp_k$ using Eq.~\ref{eq:swing}; it is hard to find closed-form equations for twist angles $\boldsymbol{\phi}$. Furthermore, estimating shape parameter $\boldsymbol{\beta}$ is non-trivial. To bypass the challenges, similarly to~\cite{li2021hybrik}, we involve the network called as twist angle and shape estimator $f^\text{TS}:\mbF_\text{img}\rightarrow[\boldsymbol{\phi}, \boldsymbol{\beta}_\text{init}]$ that estimates twist angle $\boldsymbol{\phi}=\{\phi_k\}^K_{k=1}$ and shape parameter $\boldsymbol{\beta}_\text{init}$ from image features $\mbF_\text{img}$. To resolve the discontinuity issue, it directly estimates cosine value $c_{\phi_k}$ and sine value $s_{\phi_k}$ of twist angle instead of estimating $\phi_k$. The axis of twist rotation is $\mbt_k$ and thus, twist rotation matrix $\mbR^\text{tw}_k$ can be derived by the axis $\mbt_k$ and the angle $\boldsymbol{\phi}$ as follows:
\begin{eqnarray}
    \mbR^\text{tw}_k = \textbf{I} + \frac{\sin{\phi}_k}{\|\mbt_k\|}[\mbt_k]_\times + \frac{(1-\cos{\phi}_k)}{\|\mbt_k\|^2}[\mbt_k]^2_\times,
\end{eqnarray}
where $[\mbt_k]_\times$ is the skew-symmetric matrix of $\mbt_k$. 

Finally, the relative rotation matrix $\mbR$ can be determined as follows:
\begin{eqnarray}
    \mbR = \mbR^\text{sw}\mbR^\text{tw}.
\label{eq:final rotation matrix}
\end{eqnarray}
After obtaining the rotation matrix $\mbR$, we convert it to 6D rotation representation and obtain the pose parameters $\boldsymbol{\theta}_\text{init}$. We initialize the camera parameter $\mbC_\text{init}$ as [0.9, 0, 0] and use the constant values during the inverse kinematics step.

\subsection{3D Mesh Refinement via Relation-Aware Transformer}
\label{method:Refinement}

The relation-aware refiner $f^\text{Ref}:[\mbF_\text{img}, \boldsymbol{\Theta_\text{init}}]\rightarrow\boldsymbol{\Theta_\text{ref}}$ is proposed to refine the initial SMPL parameters based on the vision Transformer architecture~\cite{dosovitskiy2020image}. Its input is the concatenation of image features $\mbF_\text{img}$ and SMPL parameters $\boldsymbol{\Theta}_\text{init}=[\boldsymbol{\theta}_\text{init}; \boldsymbol{\beta}_\text{init}; \mbC_\text{init}]$ which are obtained from the IK process. We use $N \times K$ as the sequence length of the Transformer where $N$ is the maximum number of people for the input and $K$ is the number of joints for one person. By rearranging and concatenating image features $\mbF_\text{img}$ with $\boldsymbol{\Theta}_\text{init}$, we generate the $(N\times K)\times2,067$ array as the input to the Transformer (see supplemental for details). We obtain the $\Delta \boldsymbol{\Theta}_\text{ref}$ from the Transformer and the final SMPL parameter is obtained as follows:
\begin{eqnarray}
    \boldsymbol{\Theta}_\text{ref} = \boldsymbol{\Theta}_\text{init} + \Delta \boldsymbol{\Theta}_\text{ref}.
\end{eqnarray}

From the refined parameter $\boldsymbol{\Theta}_\text{ref}=[\boldsymbol{\theta}_\text{ref};\boldsymbol{\beta}_\text{ref}; \mbC_\text{ref}]$, 3D meshes $\mbM$ are obtained, and corresponding 3D skeletons $\mbP_\text{ref}$ are further obtained by applying the mesh-to-joint regressor~\cite{smpl} to mesh vertices. 

When constituting the Transformer, we use the masking input patch as METRO~\cite{lin2021end}: randomly 0 to 30\% of input patches are masked and this makes the Transformer learn non-local interactions. We select not to use the positional embedding while using the masking scheme from results in Table~\ref{tab:ablation study}.

\noindent \textbf{Sampling interacting persons.}\label{method:choose-person} The number of persons $M$ varies depending on the image $\mbI$; while the relation-aware refiner $f^\text{Ref}$ requires to fix $N$ which is the maximum number of persons in the input. We set $N$ as $3$ according to the ablation study shown in Table~\ref{tab:ablation study}. For images having less than $N$ persons ($M<N$), we apply Transformer once by simply zero-padding unoccupied inputs, while for images having more than $N$ persons ($M>N$), we need to apply Transformers multiple times by sampling the interacting persons. The sampling scheme during training and testing stages are proposed as follows: At training, we randomly sample multiple data consisting of $N$ persons so that Transformer can see various combinations as epochs go. At testing, we run $f^\text{Ref}$ exactly $M$ times, getting results once for each person. At each run, we set each person as the target to refine, inputting $N-1$ closest persons as contexts.

\subsection{Training Method}

We use PyTorch to implement our pipeline. A single NVIDIA TITAN GPU is used for each experiment with a batch size of 64. The Adam optimizer~\cite{kingma2014adam} is used for the optimization with a learning rate of $5 \times 10^{-5}$ for relation-aware Transformer and $1 \times 10^{-4}$ for all other networks, respectively. We decrease the learning rate exponentially by a factor of 0.9 per each epoch. Total 100 epochs are executed for completely training our network.

To train the proposed initial 3D skeleton estimation network $f^\text{P}$, twist angle and shape estimation network $f^\text{TS}$ and relation-aware refiner $f^\text{Ref}$, we used the loss $L$ defined as follows:
\begin{eqnarray}
    L(f^\text{P}, f^\text{TS}, f^\text{Ref})= L_\text{P}(f^\text{P}) + L_\text{TS}(f^\text{TS}) + L_\text{Ref}(f^\text{Ref}).
\end{eqnarray}
Each loss term is detailed in the remainder of this subsection.

\noindent \textbf{Skeleton loss $L_\text{P}$.} We use multiple $L1$ losses using 2D and 3D skeletons in absolute and relative coordinate spaces to train the initial skeleton estimation network $f^\text{P}$ using the loss $L_\text{P}$ as follows:
\begin{eqnarray}
    L_\text{P}(f^P) &=& \| \mbP - \hat{\mbP}^\text{3D}_\text{abs} \|_1 + \| \mbP_\text{rel} - \hat{\mbP}^\text{3D}_\text{rel} \|_1\nonumber\\ &+& \|\mbP_\text{img} - \hat{\mbP}^\text{2D} \|_1 + \| \Pi(\mbP_\text{rel}) - \hat{\mbP}^\text{2D} \|_1
\end{eqnarray}
where $\hat{\mbP}^\text{3D}_\text{abs}$, $\hat{\mbP}^\text{3D}_\text{rel}$ and $\hat{\mbP}^\text{2D}$ are ground-truth absolute 3D skeletons, relative 3D skeletons and 2D skeletons, respectively. $\Pi$ is an orthographic projection.

\noindent \textbf{Twist angle and shape loss $L_\text{TS}$.} We use the loss $L_\text{TS}$ to train the twist angle and shape estimator $f^\text{TS}$ as follows:
\begin{eqnarray}
    L_\text{TS}(f^\text{TS}) = L_\text{angle}(f^\text{TS}) + L_\text{shape}(f^\text{TS})
\end{eqnarray}
where
\begin{eqnarray}
    L_\text{angle}(f^\text{TS}) \;&=&\; \frac{1}{K}\sum^K_{k=1}\|(c_{\phi_k}, s_{\phi_k})-(\cos{\hat{\phi}_k}, \sin{\hat{\phi}_k})\|_2,\\
    L_\text{shape}(f^\text{TS}) \;&=&\; \|\boldsymbol{\beta}_\text{init}-\hat{\boldsymbol{\beta}}\|_2,
\end{eqnarray}
$\hat{\phi}_k$ denotes the ground-truth twist angle and $\hat{\boldsymbol{\beta}}$ denotes the ground-truth SMPL shape parameters.

\noindent \textbf{Refinement loss $L_\text{Ref}$.} We use the loss $L_\text{Ref}$ combining several losses to train our relation-aware refinement network $f^\text{Ref}$ and additional discriminators $D=\{D_{\theta}, D_{\beta}\}$ as follows:
\begin{eqnarray}
    L_\text{Ref}(f^\text{Ref}, D) = L_\text{mesh}(f^\text{Ref}) + L_\text{pose}(f^\text{Ref}) + L_\text{adv}(f^\text{Ref}) + L_\text{adv}(D)
\end{eqnarray}
where
\begin{eqnarray}
    L_\text{mesh}(f^\text{Ref}) = \| \boldsymbol{\theta}_\text{ref} - \hat{\boldsymbol{\theta}} \|_2 + \|\boldsymbol{\beta}_\text{ref} - \hat{\boldsymbol{\beta}} \|_2
\end{eqnarray}
enforces the estimated pose $\boldsymbol{\theta}_\text{ref}$ and shape $\boldsymbol{\beta}_\text{ref}$ parameters close to the ground-truth pose $\hat{\boldsymbol{\theta}}$ and shape $\hat{\boldsymbol{\beta}}$ parameters of the meshes,
\begin{eqnarray}
    L_\text{pose}(f^\text{Ref})=\| \mbP_\text{ref} - \hat{\mbP}^\text{3D}_\text{rel} \|^2_2 + \| \Pi(\mbP_\text{ref}) - \hat{\mbP}^\text{2D} \|^2_2
\end{eqnarray}
enforces estimated 3D skeletons $\mbP_\text{ref}$ and its orthographic projection $\Pi(\mbP_\text{ref})$ close to ground-truth 3D and 2D skeletons ($\hat{\mbP}^\text{3D}_\text{rel}$, $\hat{\mbP}^\text{2D}$), respectively,
\begin{eqnarray}
    L_\text{adv}(D) &=& \| D_{\theta}(\boldsymbol{\theta}_\text{ref}) - 0 \|_2 + \| D_{\theta}(\boldsymbol{\theta}_\text{real}) - 1 \|_2 \nonumber\\
    &+& \| D_{\beta}(\boldsymbol{\beta}_\text{ref}) - 0 \|_2 + \| D_{\beta}(\boldsymbol{\beta}_\text{real}) - 1 \|_2
\end{eqnarray}
trains discriminators $D_{\theta}$, $D_{\beta}$ to classify real SMPL parameter $\boldsymbol{\theta}_\text{real}$ and $\boldsymbol{\beta}_\text{real}$ as real (i.e. $1$) and estimated SMPL parameter $\boldsymbol{\theta}_\text{ref}$ and $\boldsymbol{\beta}_\text{ref}$ as fake (i.e. $0$) and 
\begin{eqnarray}
    L_\text{adv}(f^\text{Ref}) = \| D_{\theta}(\boldsymbol{\theta}_\text{ref}) - 1 \|_2+\| D_{\beta}(\boldsymbol{\beta}_\text{ref}) - 1 \|_2
\end{eqnarray}
enforces the estimated $\boldsymbol{\theta}_\text{ref}$ and $\boldsymbol{\beta}_\text{ref}$ become realistic to deceive two discriminators $D_{\theta}$ and $D_{\beta}$ to say that it is the real sample (i.e. $1$).

\section{Experiments}

\noindent \textbf{Setup.} We involved multiple datasets to train our model. We used Human3.6M \cite{ionescu2013human3}, MPI-INF-3DHP~\cite{mehta2017monocular}, LPS~\cite{johnson2010clustered}, MSCOCO~\cite{lin2014microsoft}, MPII~\cite{andriluka20142D} datasets as the training data, which are the same setting as Kanazawa et al.~\cite{kanazawa2018end}. Additionally, MuCo-3DHP~\cite{mehta2018single}, CMU-Panoptic~\cite{joo2015panoptic}, SAIL-VOS~\cite{hu2019sail}, SURREAL~\cite{varol2017learning}, AIST++~\cite{li2021learn} are used to calculate the $L_\text{P}(f^\text{P})$. For evaluation, we used 3DPW \cite{von2018recovering}, MuPoTs~\cite{mehta2018single}, and AGORA~\cite{patel2021Agora}: The 3DPW dataset is an outdoor 3D human pose benchmark involving real sequences. It contains diverse subjects, various backgrounds and occlusion scenario. The MuPoTS dataset contains both real indoor and outdoor sequences having multiple persons occluding each other. AGORA is the synthetic benchmark having multi-person within it. Image contains many persons with various clothes, ages and ethnicities.

\begin{table}[t]
\captionsetup{font=scriptsize}
\begin{minipage}[t]{0.48\linewidth}
\centering
\caption{SOTA comparisons on 3DPW.}
\label{tab:comparison on 3DPW}
\resizebox{\textwidth}{!}{
\begin{tabular}{lccc} 
\hline 
\textbf{Method} & MPJPE($\downarrow$) & PA-MPJPE($\downarrow$) & PVE($\downarrow$) \\
\hline
HMR \cite{kanazawa2018end} & 130.0  & 76.7 & - \\
temporal HMR \cite{kanazawa2019learning} & 116.5  & 72.6 & 139.3 \\
BMP \cite{zhang2021body} & 104.1 & 63.8 & 119.3 \\
SPIN \cite{kolotouros2019learning} & 96.6  & 59.2 & 116.4 \\
VIBE \cite{kocabas2020vibe} & 93.5  & 56.5 & 116.4 \\
ROMP(Resnet-50) \cite{sun2021monocular} & 89.3  & 53.5 & 105.6 \\
ROMP(HRNet-32) \cite{sun2021monocular} & 85.5  & 53.3 & 103.1 \\
PARE(Resnet-50) \cite{kocabas2021pare} & 82.9  & 52.3 & 99.7 \\
PARE(HRNet-50) \cite{kocabas2021pare} & 82.0  & 50.9 & 97.9 \\
3DCrowdNet \cite{choi20213Dcrowdnet} & 82.8 & 52.2 & 100.2 \\
HybrIK~\cite{li2021hybrik} & 80.0 & 48.8 & 94.5 \\
METRO \cite{lin2021end} & 77.1 & 47.9 & 88.2 \\
MeshLeTemp \cite{tran2022meshletemp} & 74.8 & 46.8 & 86.5 \\
Mesh Graphormer \cite{lin2021mesh} & 74.7 & 45.6 & 87.7 \\
\hline
Ours & \textbf{66.0} & \textbf{39.0} & \textbf{76.3} \\
\hline
\end{tabular}}
\end{minipage}
\begin{minipage}[t]{0.48\linewidth}
\centering
\caption{SOTA Comparisons on AGORA.}
\label{tab:comparison on Agora}
\setlength{\tabcolsep}{3pt}
\renewcommand{\arraystretch}{1.435}
\resizebox{\textwidth}{!}{
\begin{tabular}{l|cc|cc}
\hline
\multirow{2}{*}{\textbf{Method}} & \multicolumn{2}{c}{All} & \multicolumn{2}{|c}{Matched}\\\cline{2-5} 
 & NMVE($\downarrow$) & NMJE($\downarrow$) & MVE($\downarrow$) & MPJPE($\downarrow$) \\
\hline
ROMP \cite{sun2021monocular} & 227.3  & 236.6 & 161.4 & 168.0\\
HMR \cite{kanazawa2018end} & 217.0  & 226.0 & 173.6 & 180.5\\
SPIN \cite{kolotouros2019learning} & 216.3  & 223.1 & 168.7 & 175.1\\
PyMAF \cite{zhang2021pymaf} & 200.2  & 207.4 & 168.2 & 174.2\\
EFT \cite{joo2021exemplar} & 196.3  & 203.6 & 159.0 & 165.4\\
HybrIK \cite{li2021hybrik} & - & 188.5 & - & 156.1 \\
PARE \cite{kocabas2021pare} & 167.7  & 174.0 & 140.9 & 146.2\\
SPEC \cite{kocabas2021spec} & 126.8 & 133.7 & 106.5 & 112.3 \\
\hline
Ours & \textbf{104.5} & \textbf{110.4} & \textbf{86.7} & \textbf{91.6} \\
\hline
\end{tabular}}
\end{minipage}
\end{table}

\begin{table}[t]
\captionsetup{font=scriptsize}
\centering
\caption{3DPCK relevant on MuPoTS-3D dataset for all sequences. The above table shows accuracy only for all groundtruths. The below table shows accuracy only for matched groundtruths.}
\label{tab:comparison on MuPoTS}
\resizebox{\columnwidth}{!}{
\begin{tabular}{lccccccccccccccccccccccc}
\hline
\textbf{Method}-3DPCK($\uparrow$) & S1 & S2 & S3 & S4 & S5 & S6 & S7 & S8 & S9 & S10 & S11 & S12 & S13 & S14 & S15 & S16 & S17 & S18 & S19 & S20 & Avg \\
\hline
\multicolumn{12}{l}{\textbf{Accuracy for all groundtruths}}\\
Jiang et al. \cite{jiang2020coherent} & - & - & - & - & - & - & - & - & - & - & - & - & - & - & - & - & - & - & - & - & 69.1 \\
ROMP \cite{sun2021monocular} & 89.8 & 73.1 & 67.2 & 68.4 & 78.9 & 41.0 & 68.7 & 68.2 & 70.1 & 85.4 & 69.2 & 63.2 & 66.5 & 60.9 & 78.1 & 77.4 & 75.1 & 80.7 & 74.0 & 61.1 & 71.9  \\
SPEC \cite{kocabas2021spec} & 87.2 & 69.4 & 69.0 & 71.5 & 78.5 & 63.8 & 69.1 & 66.2 & 71.5 & 85.7 & 69.2 & 63.2 & 66.5 & 60.9 & 78.1 & 77.4 & 75.1 & 80.7 & 74.0 & 61.1 & 71.9 \\
BMP \cite{zhang2021body} & - & - & - & - & - & - & - & - & - & - & - & - & - & - & - & - & - & - & - & - & 73.8 \\
PARE \cite{kocabas2021pare} & 87.7 & 65.4 & 66.4 & 67.7 & 81.9 & 62.5 & 64.9 & 69.9 & 73.8 & 88.5 & 80.1 & 79.2 & 74.5 & 62.9 & 81.6 & 84.5 & 89.6 & 83.7 & 73.7 & 66.5 & 75.3  \\
Moon et al. \cite{moon2019camera} & 94.4 & 77.5 & 79.0 & 81.9 & 85.3 & 72.8 & 81.9 & 75.7 & \textbf{90.2} & 90.4 & 79.2 & 79.9 & 75.1 & 72.7 & 81.1 & 89.9 & 89.6 & 81.8 & 81.7 & 76.2 & 81.8 \\
Metrabs \cite{sarandi2020metrabs} & 94.0 & 82.6 & 88.4 & 86.5 & 87.3 & 76.2 & 85.9 & 66.9 & 85.8 & 92.9 & 81.8 & 89.9 & 77.6 & 68.5 & 85.6 & 92.3 & 89.3 & 85.1 & 78.2 & 71.6 & 83.3 \\
Cheng et al. \cite{cheng2021monocular} & 93.4 & \textbf{91.3} & 84.7 & 83.3 & 89.1 & 85.2 &  \textbf{95.4} & \textbf{92.1} & 89.5 & 93.1 & 85.4 & 85.7 & \textbf{89.9} & \textbf{90.1} & 88.8 & 93.7 & 92.2 & 87.9 & \textbf{89.7} & \textbf{91.9} & 89.6 \\
\hline
Ours & \textbf{97.3} & 84.7 & \textbf{91.1} & \textbf{89.9} & \textbf{92.9} & \textbf{89.8} & 92.2 & 87.1 & 89.1 & \textbf{94.0} & \textbf{88.6} & \textbf{92.9} & 84.6 & 80.4 & \textbf{94.3} & \textbf{96.7} & \textbf{98.8} & \textbf{91.5} & 86.1 & 76.7 & \textbf{89.9} \\
\hline
\multicolumn{12}{l}{\textbf{Accuracy only for matched groundtruths.}}\\
Jiang et al. \cite{jiang2020coherent} & - & - & - & - & - & - & - & - & - & - & - & - & - & - & - & - & - & - & - & - & 72.2 \\
ROMP \cite{sun2021monocular} & 92.1 & 81.9 & 69.8 & 69.1 & 85.9 & 43.2 & 69.3 & 70.7 & 70.1 & 85.4 & 69.2 & 63.2 & 68.0 & 63.6 & 78.1 & 77.6 & 75.4 & 80.7 & 74.5 & 72.9 & 74.0 \\
SPEC \cite{kocabas2021spec} & 88.1 & 78.5 & 69.6 & 71.6 & 81.0 & 63.8 & 69.1 & 77.4 & 71.5 & 85.7 & 69.2 & 63.2 & 68.0 & 63.6 & 78.1 & 77.6 & 75.4 & 80.7 & 74.5 & 72.9 & 74.0 \\
BMP \cite{zhang2021body} & - & - & - & - & - & - & - & - & - & - & - & - & - & - & - & - & - & - & - & - & 75.3 \\
PARE \cite{kocabas2021pare} & 88.8 & 77.1 & 66.7 & 67.7 & 83.2 & 62.5 & 64.9 & 77.9 & 73.8 & 88.5 & 80.1 & 79.2 & 76.4 & 65.8 & 81.6 & 85.4 & 89.6 & 83.7 & 74.2 & 82.5 & 77.5  \\
Moon et al. \cite{moon2019camera} & 94.4 & 78.6 & 79.0 & 82.1 & 86.6 & 72.8 & 81.9 & 75.8 & \textbf{90.2} & 90.4 & 79.4 & 79.9 & 75.3 & 81.0 & 81.0 & 90.7 & 89.6 & 83.1 & 81.7 & 77.3 & 82.5  \\
Metrabs \cite{sarandi2020metrabs} & 94.0 & 86.5 & 89.0 & 87.1 & 91.1 & 77.4 & 90.2 & 75.7 & 85.8 & 92.9 & 86.0 & 90.7 & 83.8 & 82.0 & 85.6 & 94.3 & 89.8 & 89.6 & \textbf{86.5} & \textbf{91.7} & 87.5 \\
Cheng et al. \cite{cheng2021monocular} & - & - & - & - & - & - & - & - & - & - & - & - & - & - & - & - & - & - & - & - & 89.6 \\
\hline
Ours & \textbf{97.6} & \textbf{94.1} & \textbf{90.7} & \textbf{89.6} & \textbf{95.2} & \textbf{88.7} & \textbf{94.2} & \textbf{88.4} & 89.2 & \textbf{93.7} & \textbf{89.1} & \textbf{93.2} & \textbf{86.6} & \textbf{90.5} & \textbf{94.4} & \textbf{97.4} & \textbf{98.5} & \textbf{91.9} & 86.1 & 84.8 & \textbf{91.7}  \\
\hline
\end{tabular}
}
\end{table}

\begin{table}[t]
\captionsetup{font=scriptsize}
\centering
\caption{Ablation study of the effectiveness of IK, refiner, positional embedding, masking input patch and comparison between the different number of Transformer's input persons on 3DPW.}
\label{tab:ablation study}
\resizebox{\textwidth}{!}{
\begin{tabular}{lccc|lccc} 
\hline 
\textbf{Method} & MPJPE($\downarrow$) & PA-MPJPE($\downarrow$) & MVE($\downarrow$) & \textbf{Method} & MPJPE($\downarrow$) & PA-MPJPE($\downarrow$) & MVE($\downarrow$) \\
\hline
Ours w/o IK, w/o Ref & 71.8 & 42.5 & - & Ours (N=1) & 66.9 & 39.4 & 77.0 \\
Ours w/o Ref & 67.3 & 39.3 & 77.5 & Ours (N=2) & 66.7 & 39.3 & 76.7 \\
Ours w/ positional embedding & 68.3 & 39.4 & 78.3 & Ours (N=3) & \textbf{66.0} & \textbf{39.0} & \textbf{76.3} \\
Ours w/o masking input patch & 67.1 & 39.0 & 76.6 & Ours (N=4) & 66.6 & 39.4 & 77.2 \\
\hline
\end{tabular}}
\end{table}

\begin{figure*}[ht!]
\centering
\includegraphics[width=1\textwidth]{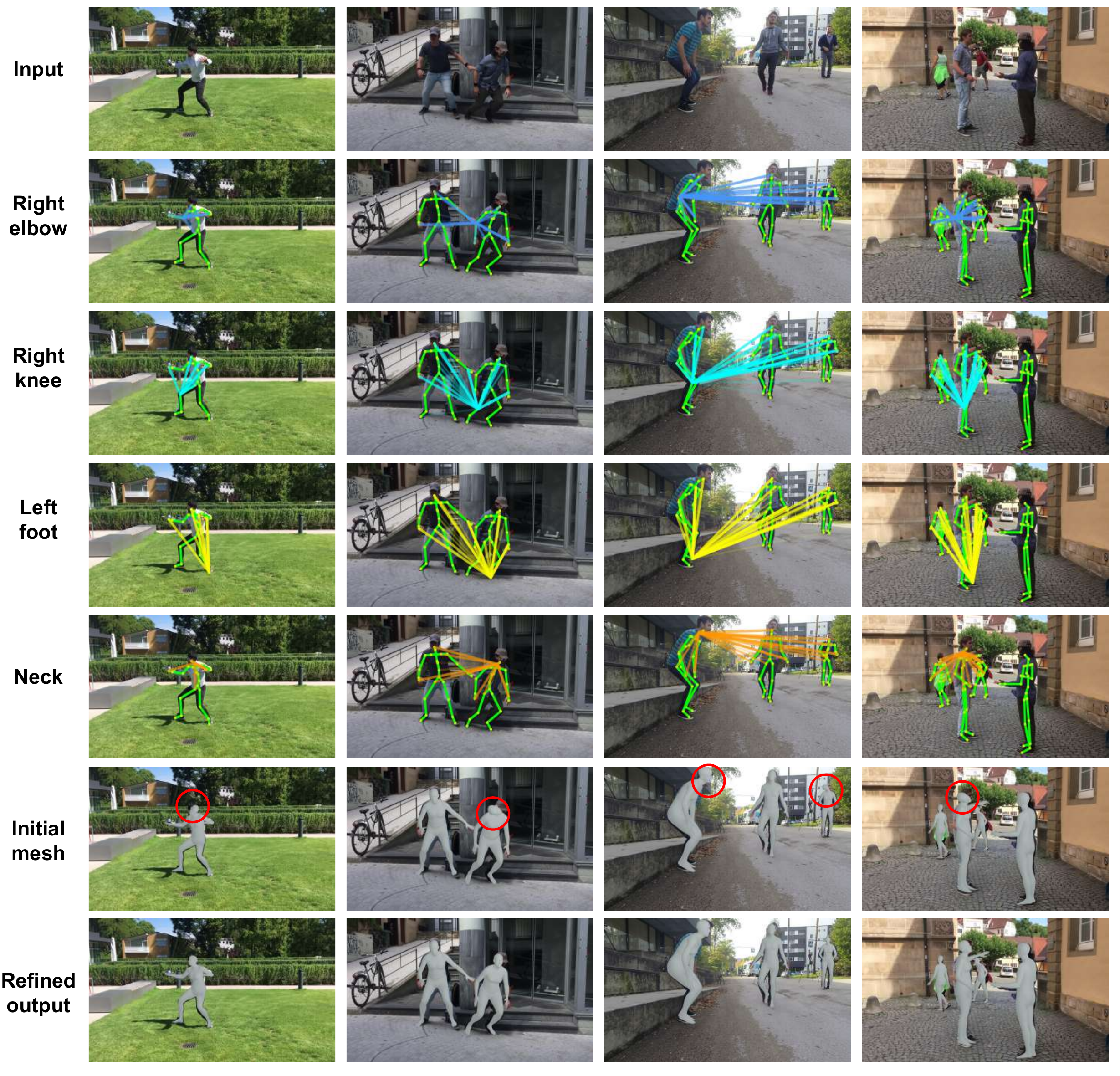}
\caption{Attention visualization: (Row 1) input image, (Row 2-5) part-based attentions obtained within an intra-person in the column 1, those among two persons in the column 2 and those among three persons in the column 3-4, (Row 6) initial mesh obtained from inverse kinematics, (Row 7) refined mesh after mesh refinement module. We visualize the self-attentions between a specified joint and all other joints, where brighter color and thicker line indicate stronger attentions.}
\label{attention}
\end{figure*}

\noindent \textbf{Measures.} For 3DPW dataset, we measure the accuracy of ours with widely used evaluation metrics to compare with others: MPJPE, PA-MPJPE and MVE: The MPJPE is the mean per joint position error which is calculated based on the Euclidean distance between ground-truth and estimated joint positions. For this, pelvis joint is aligned by moving the estimated pelvis joint to the ground-truth pelvis joint. The PA-MPJPE is Procrustes Aligned MPJPE which is calculated similarly to MPJPE; however it is measured after rigidly aligning the estimated joints to ground-truth joints via Procrustes Analysis~\cite{gower1975generalized}. The MVE is the mean vertex error which is calculated as the Euclidean distance between ground-truth and estimated mesh vertices. For MuPoTS dataset, we measure the performance of our methods using 3DPCK. The 3DPCK is the 3D percentage of correct keypoints. It counts the keypoints as correct when the Euclidean distance between the estimated joint position and its ground-truth is within a threshold. We used 150mm as the threshold following~\cite{moon2019camera}. For AGORA dataset, we measure the performance of our methods on AGORA using MPJPE, MVE, NMJE and NMVE. The MPJPE and MVE are measured on matched detections. The NMJE and NMVE are normalized MPJPE and MVE by F1 score to punish misses and false alarms in the detection.

\noindent \textbf{Baselines.} In our experiments, we have involved several state-of-the-art 3D body mesh reconstruction~\cite{joo2021exemplar,kanazawa2018end,kanazawa2019learning,kocabas2020vibe,kocabas2021pare,kocabas2021spec,kolotouros2019learning,li2021hybrik,lin2021end,lin2021mesh,tran2022meshletemp,zhang2021pymaf} pipelines for single persons to compare with ours: HMR~\cite{kanazawa2018end} and SPIN~\cite{kolotouros2019learning} are the pioneering works that first tried to infer SMPL pose and shape parameters using the CNN network for a single person. Temporal HMR~\cite{kanazawa2019learning} and VIBE~\cite{kocabas2020vibe} developed to further utilize the temporal information from a video. METRO~\cite{lin2021end} used transformer architecture for non-parametric mesh reconstruction. Mesh Graphormer~\cite{lin2021mesh} combines self-attention and graph convolution network for mesh reconstruction. MeshLeTemp~\cite{tran2022meshletemp} proposed the learnable template which reflects not only vertex-vertex interactions but also the human pose and body shape. PyMAF~\cite{zhang2021pymaf} uses a pyramidal mesh alignment feedback loop to refine the mesh based on the mesh-image alignment mechanism. EFT~\cite{joo2021exemplar} trains HMR architecture with a large-scale dataset having pseudo ground-truths. SPEC~\cite{kocabas2021spec} estimates the perspective camera to accurately infer the 3D mesh coordinates. PARE~\cite{kocabas2021pare} learns the body-part-guided attention masks to be robust to occlusions. HybrIK~\cite{li2021hybrik} estimates SMPL pose and shape parameters from estimated 3D skeletons via inverse kinematics. The single-person mesh reconstruction methods take the top-down approach using bounding boxes obtained from YOLOv4~\cite{bochkovskiy2020yolov4} for AGORA dataset, and using those obtained from ground-truth 3D skeletons for MuPoTS and 3DPW datasets, respectively.

Several multi-person frameworks are also involved for the comparisons~\cite{choi20213Dcrowdnet,jiang2020coherent,sun2021monocular,zhang2021body}: Jiang et al.~\cite{jiang2020coherent}, ROMP~\cite{sun2021monocular} and BMP~\cite{zhang2021body} are bottom-up methods that estimate the multi-person SMPL pose and shape parameters at once and simultaneously localizes multi-person instances and predicts 3D body meshes in a single stage, respectively. Jiang et al.~\cite{jiang2020coherent} proposed inter-penetration loss to avoid collision and depth ordering loss for the rendering. 3DCrowdNet~\cite{choi20213Dcrowdnet} is a top-down method that proposed to concatenate image features and 2D pose heatmaps to exploit the 2D pose-guided features for better accuracy. We also involved 3D skeleton estimation approaches~\cite{cheng2021monocular,moon2019camera,sarandi2020metrabs}: Moon et al.~\cite{moon2019camera}'s work that estimates the absolute root position and root-relative 3D skeletons focusing on camera distance. Cheng et al.~\cite{cheng2021monocular}'s work that integrates top-down method and bottom-up methods for estimating better 3D skeletons. Metrabs~\cite{sarandi2020metrabs} that is robust to truncation/occlusion variations thanks to the metric-scale heatmap representation.

\noindent \textbf{Results.} We compared ours to state-of-the-art algorithms on three challenging datasets (i.e. 3DPW, MuPoTS and AGORA). The results are summarized in Tables~\ref{tab:comparison on 3DPW}, \ref{tab:comparison on Agora} and \ref{tab:comparison on MuPoTS}: From Tables~\ref{tab:comparison on 3DPW} and~\ref{tab:comparison on Agora}, we could observe that ours obtained the superior performance compared to previous mesh reconstruction works. We obtained even better performance than works exploiting temporal information~\cite{kanazawa2019learning,kocabas2020vibe} and several multi-person 3D mesh reconstruction methods~\cite{choi20213Dcrowdnet,jiang2020coherent,sun2021monocular,zhang2021body}. Also, in Table~\ref{tab:comparison on 3DPW}, we also compared ours to HybrIK~\cite{li2021hybrik} that applied the inverse kinematics process on the pre-estimated 3D skeleton results. We outperforms it by successfully extending it towards the multi-person scenario. As shown in Table~\ref{tab:comparison on MuPoTS}, we have achieved the state-of-the-art accuracy on MuPoTS. Note that the 3D skeleton estimation methods~\cite{cheng2021monocular,moon2019camera,sarandi2020metrabs} are also included in the comparison and they produced superior performance than 3D mesh estimation approaches~\cite{kocabas2021pare,kocabas2021spec,sun2021monocular,zhang2021body}. However, our method outperforms them by delivering good pose accuracy from the initial 3D skeleton estimator, while reconstructing both poses and shapes in the form of 3D meshes. 

\begin{figure*}[ht!]
\captionsetup[subfigure]{labelformat=empty}
\centering
\subfloat[]{\includegraphics[width=0.19\textwidth]{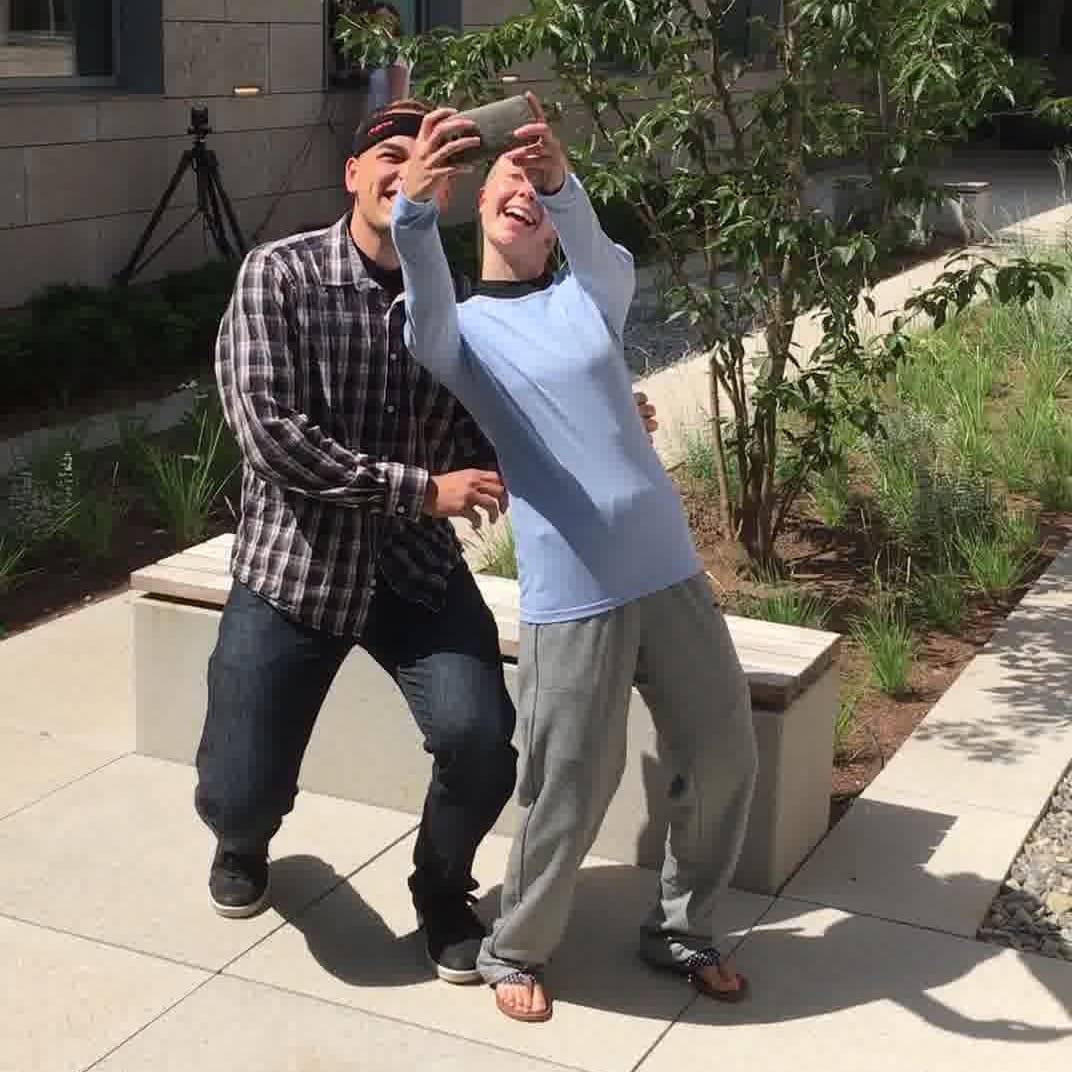}}
\subfloat[]{\includegraphics[width=0.19\textwidth]{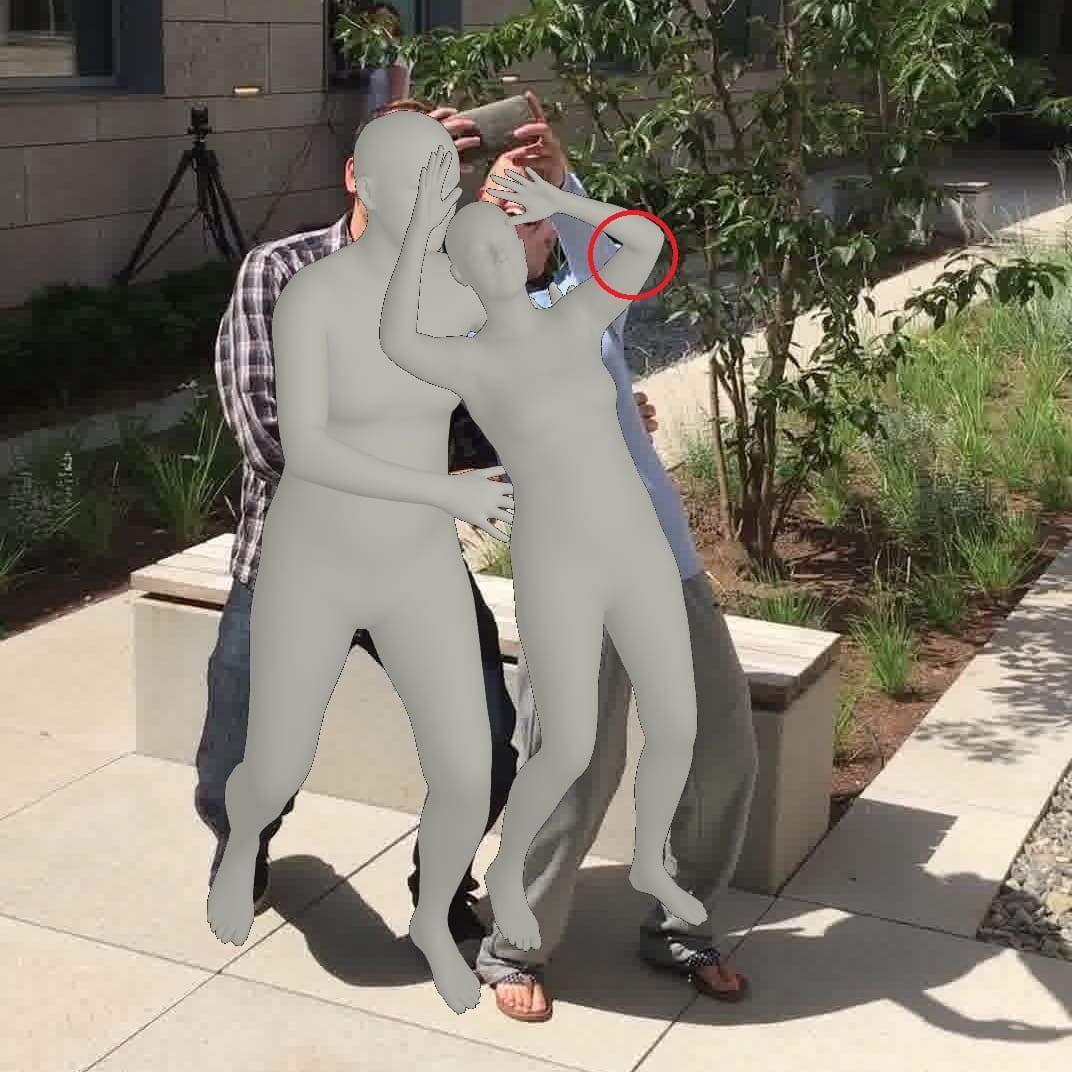}}
\subfloat[]{\includegraphics[width=0.19\textwidth]{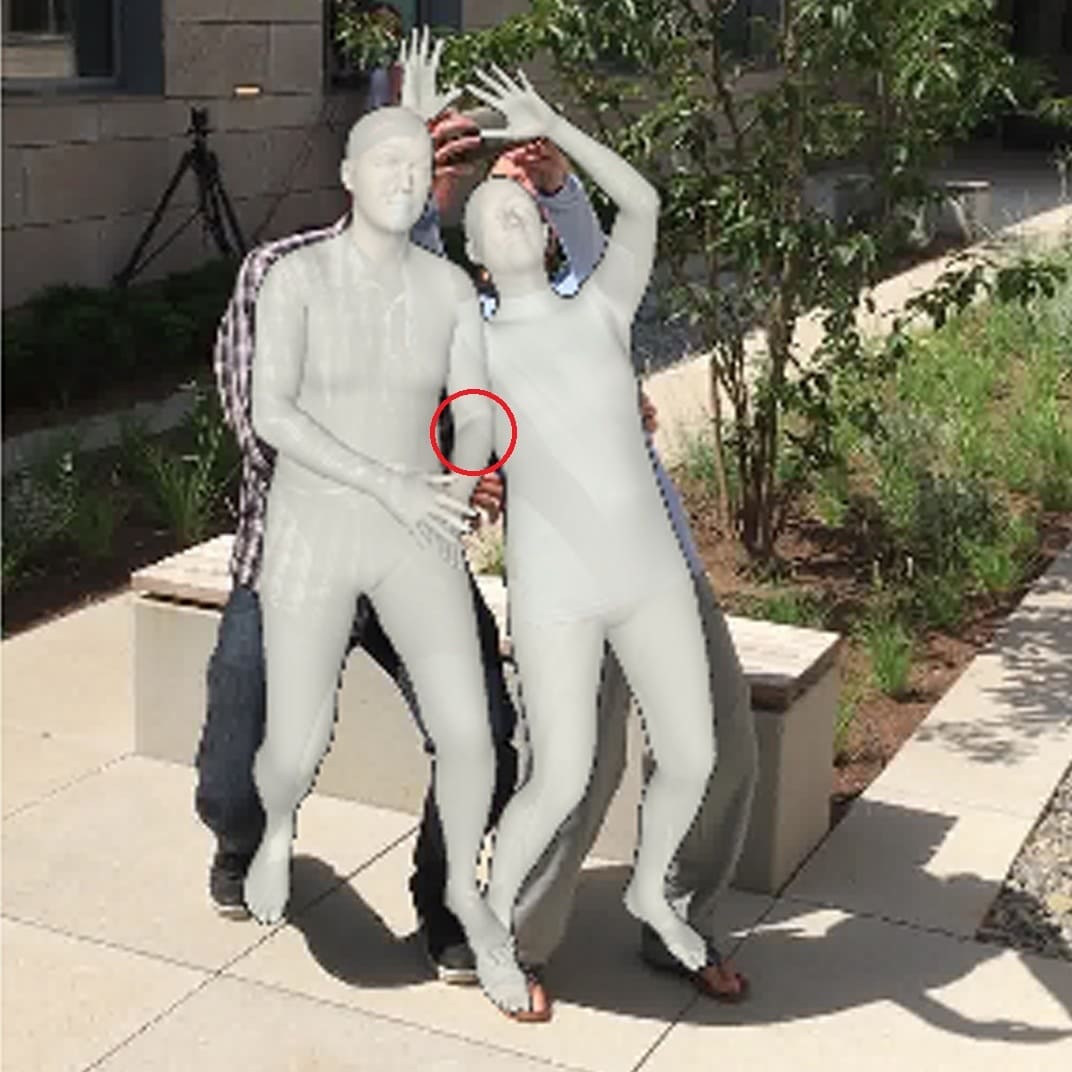}}
\subfloat[]{\includegraphics[width=0.19\textwidth]{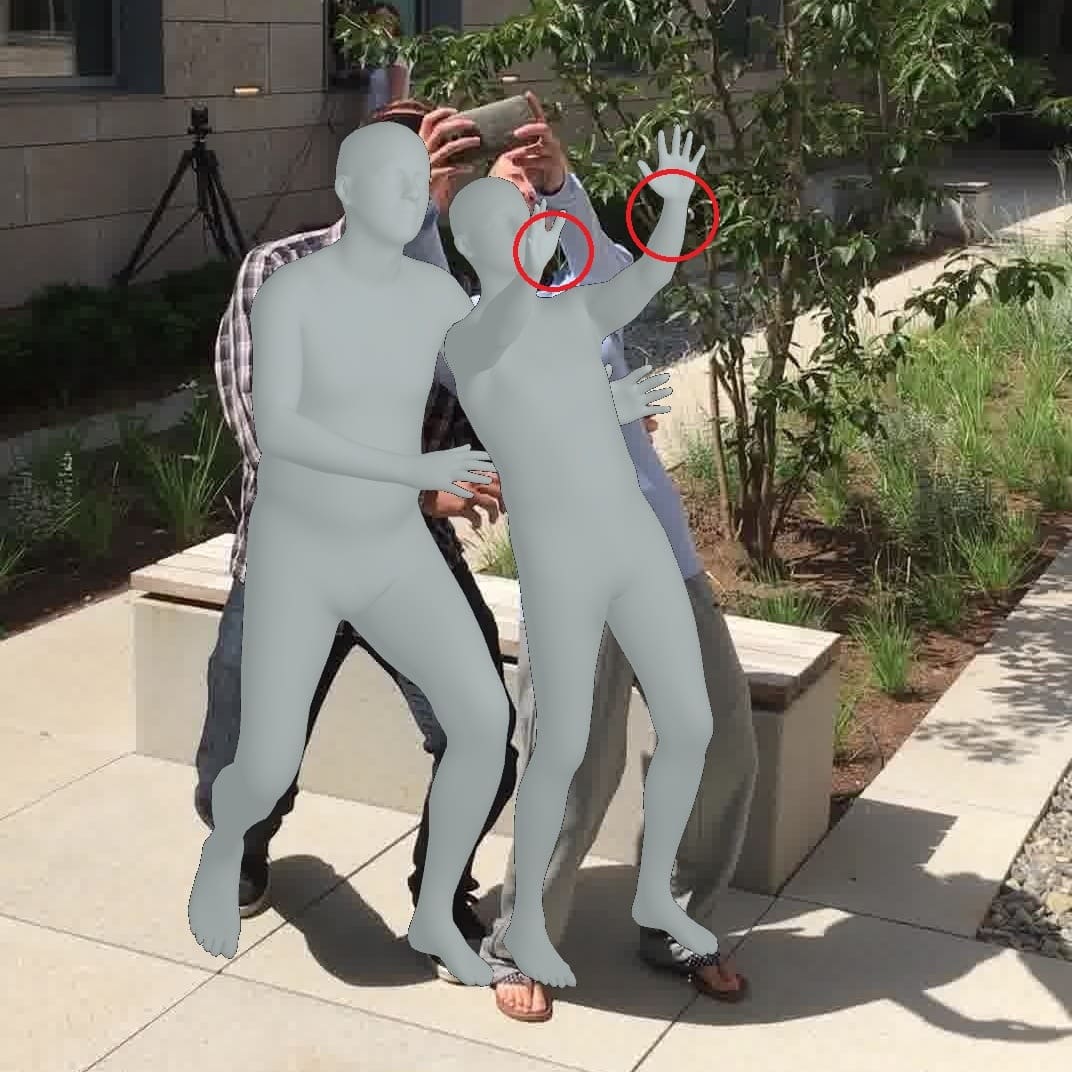}}
\subfloat[]{\includegraphics[width=0.19\textwidth]{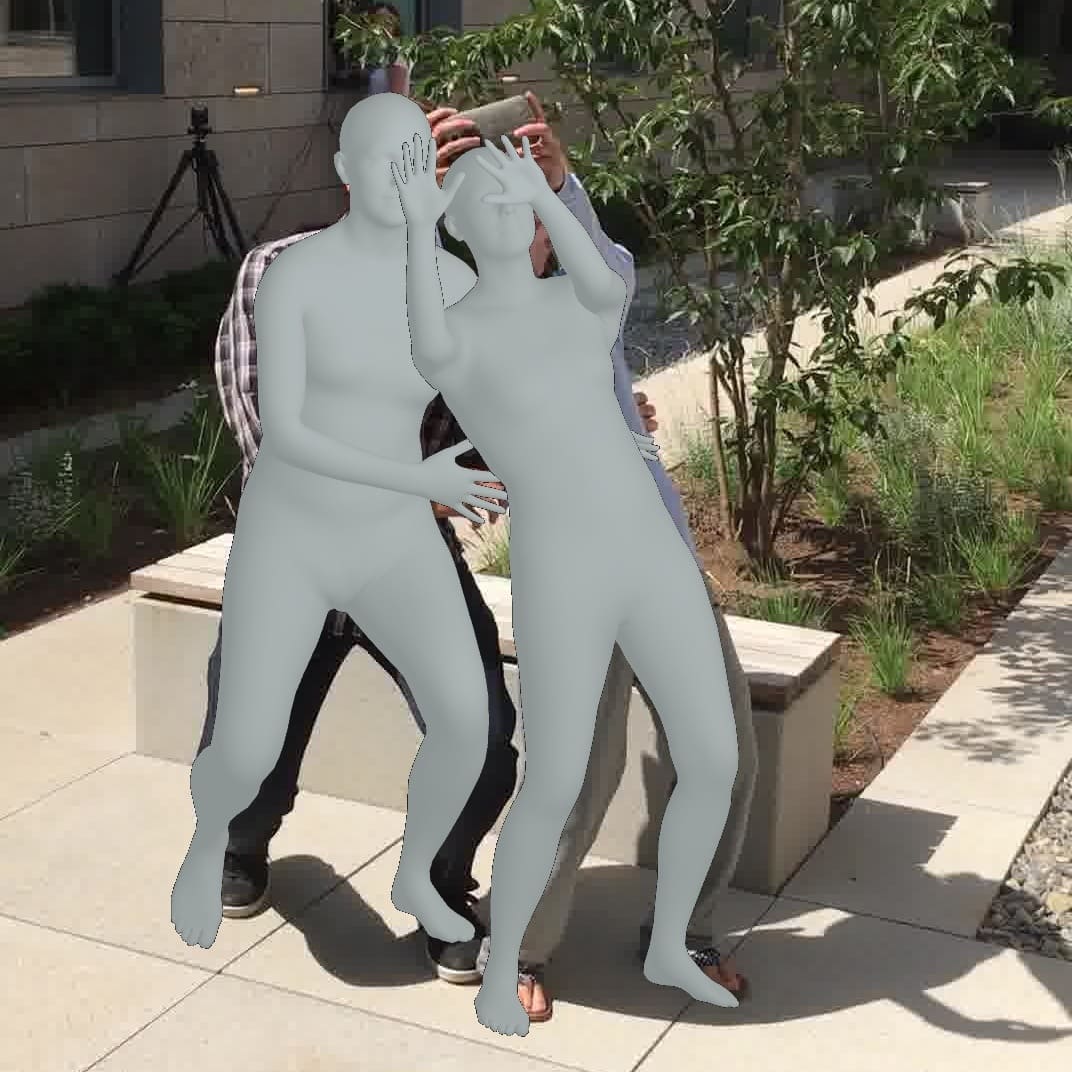}}
\\[-6.6ex]
\subfloat[]{\includegraphics[width=0.19\textwidth]{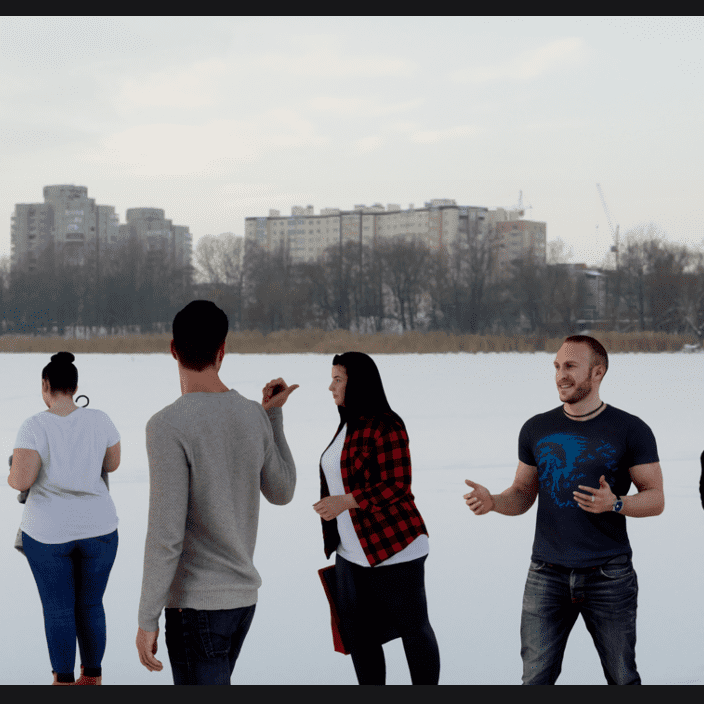}}
\subfloat[]{\includegraphics[width=0.19\textwidth]{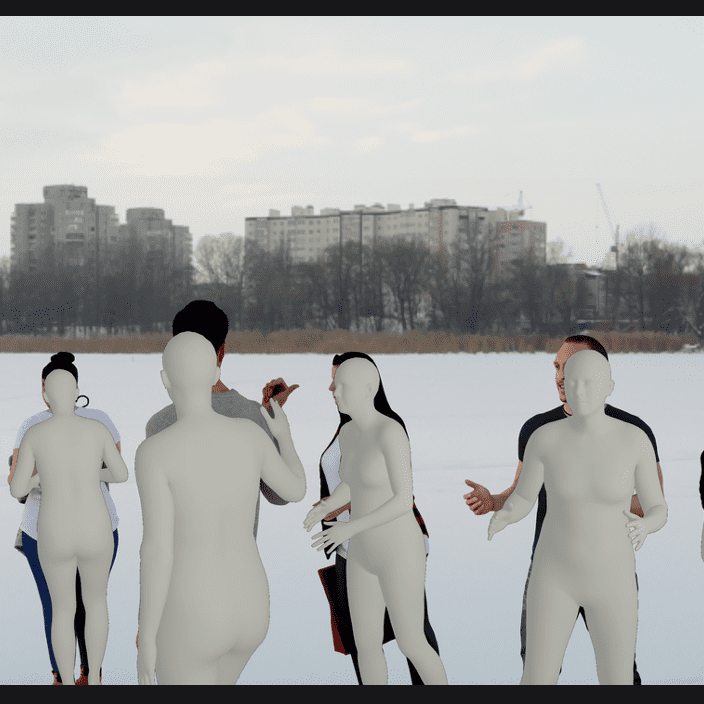}}
\subfloat[]{\includegraphics[width=0.19\textwidth]{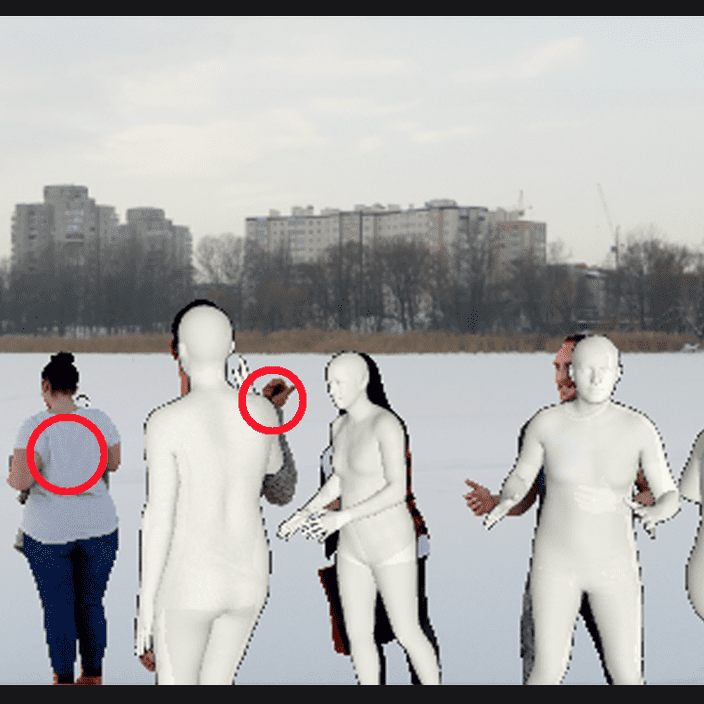}}
\subfloat[]{\includegraphics[width=0.19\textwidth]{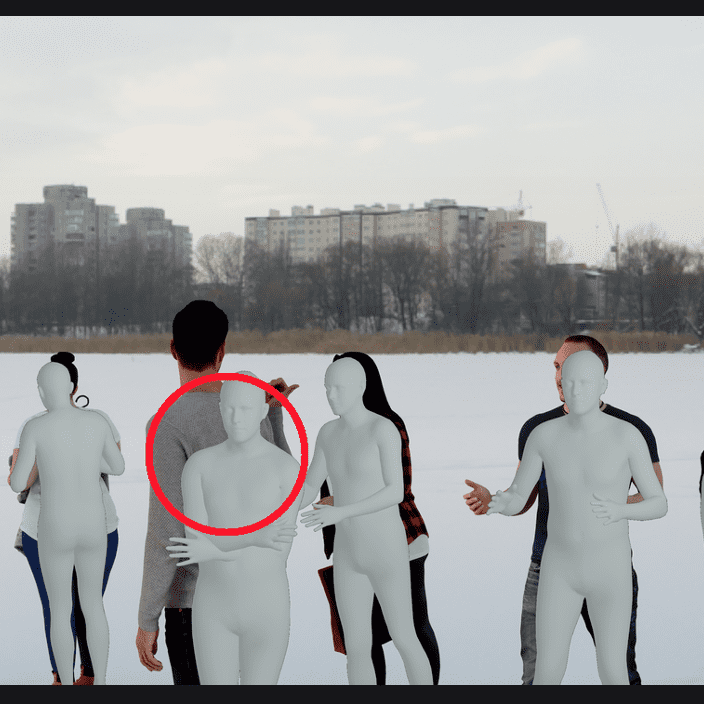}}
\subfloat[]{\includegraphics[width=0.19\textwidth]{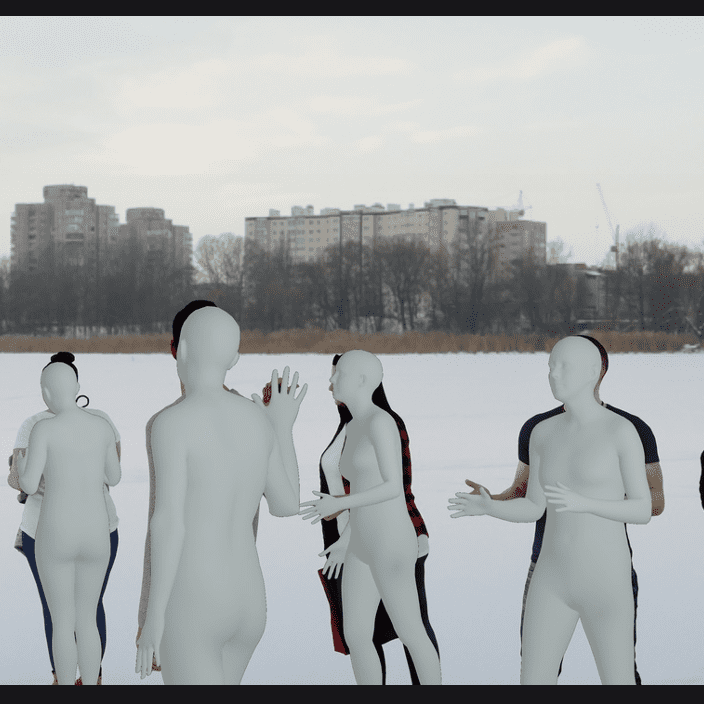}}
\\ [-6.6ex]
\subfloat[]{\includegraphics[width=0.19\textwidth]{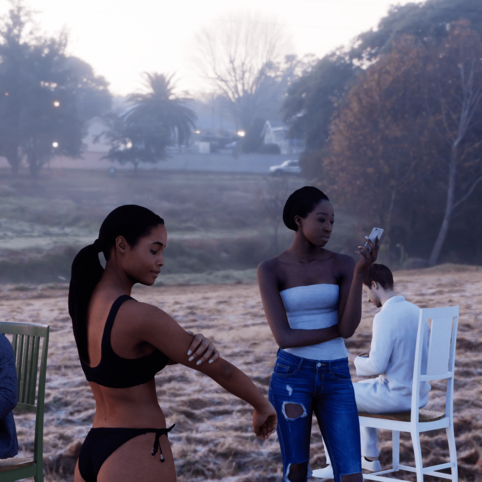}}
\subfloat[]{\includegraphics[width=0.19\textwidth]{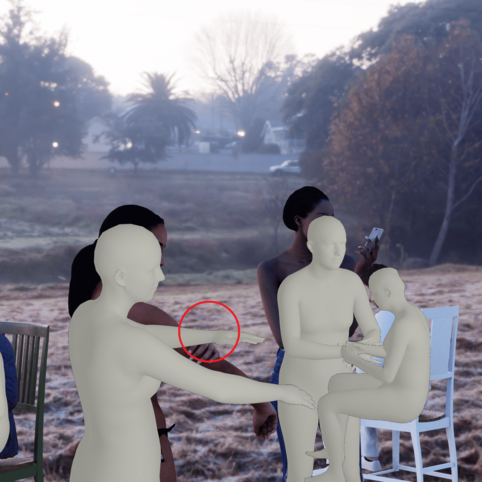}}
\subfloat[]{\includegraphics[width=0.19\textwidth]{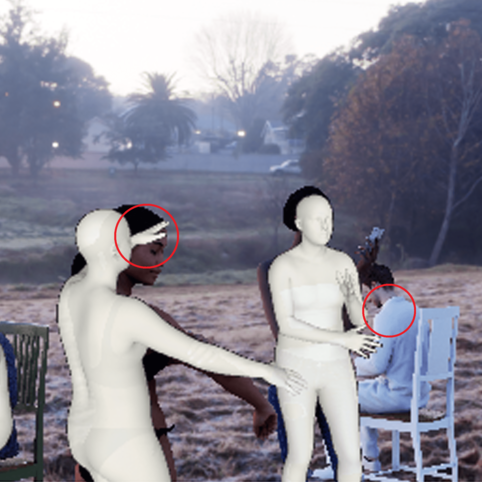}}
\subfloat[]{\includegraphics[width=0.19\textwidth]{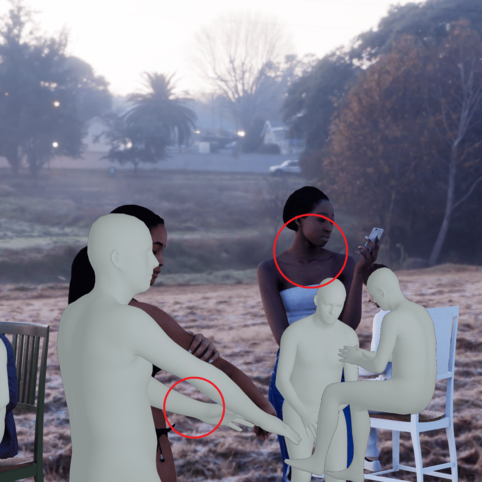}}
\subfloat[]{\includegraphics[width=0.19\textwidth]{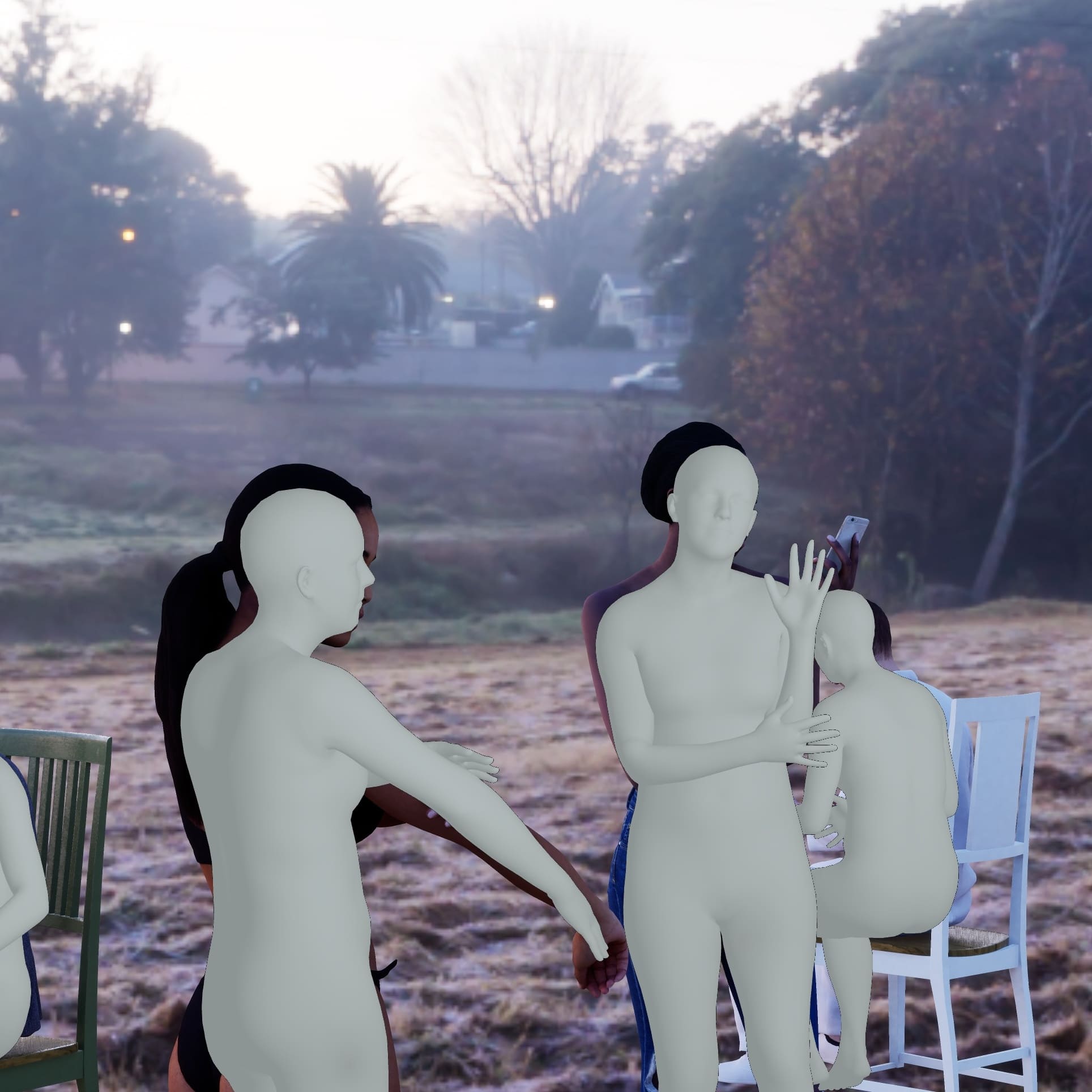}}
\\ [-6.6ex]
\subfloat[]{\includegraphics[width=0.19\textwidth]{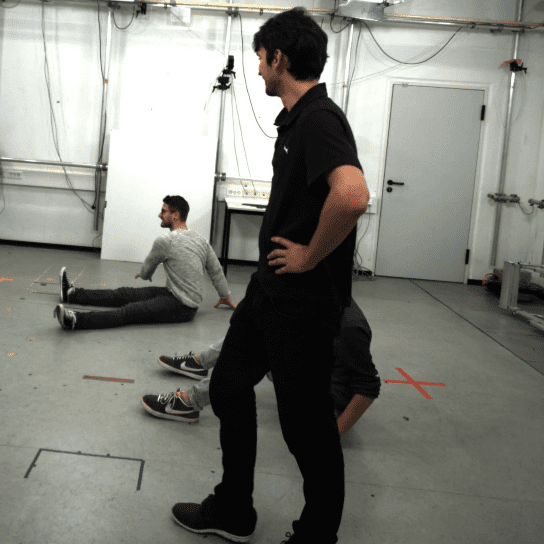}}
\subfloat[]{\includegraphics[width=0.19\textwidth]{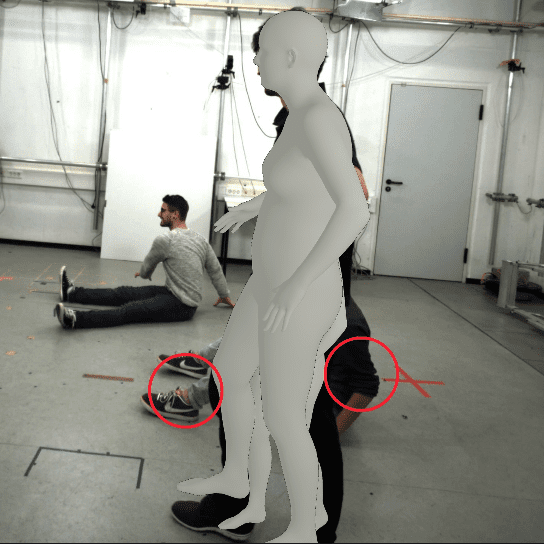}}
\subfloat[]{\includegraphics[width=0.19\textwidth]{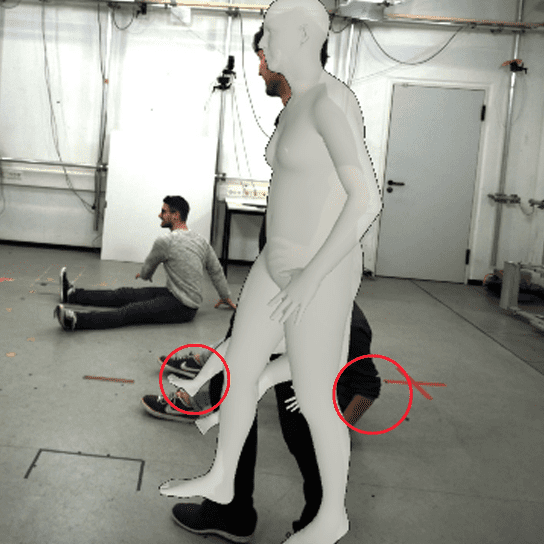}}
\subfloat[]{\includegraphics[width=0.19\textwidth]{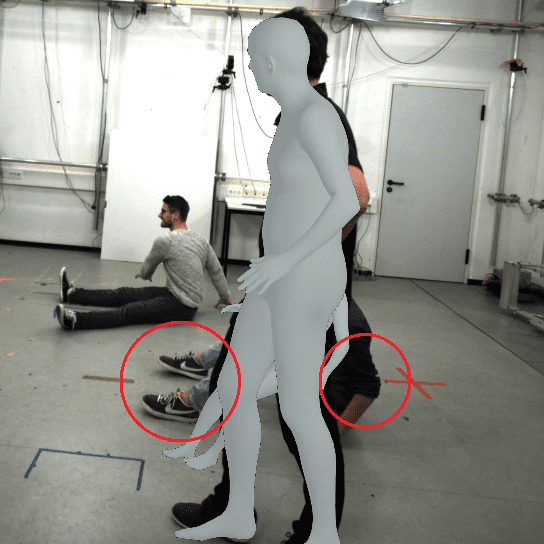}}
\subfloat[\textit{Ours}]{\includegraphics[width=0.19\textwidth]{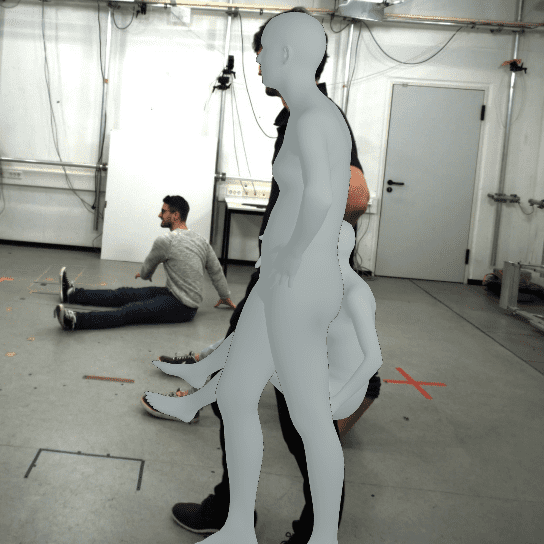}}
\\ [-6.6ex]
\subfloat[Input]{\includegraphics[width=0.19\textwidth]{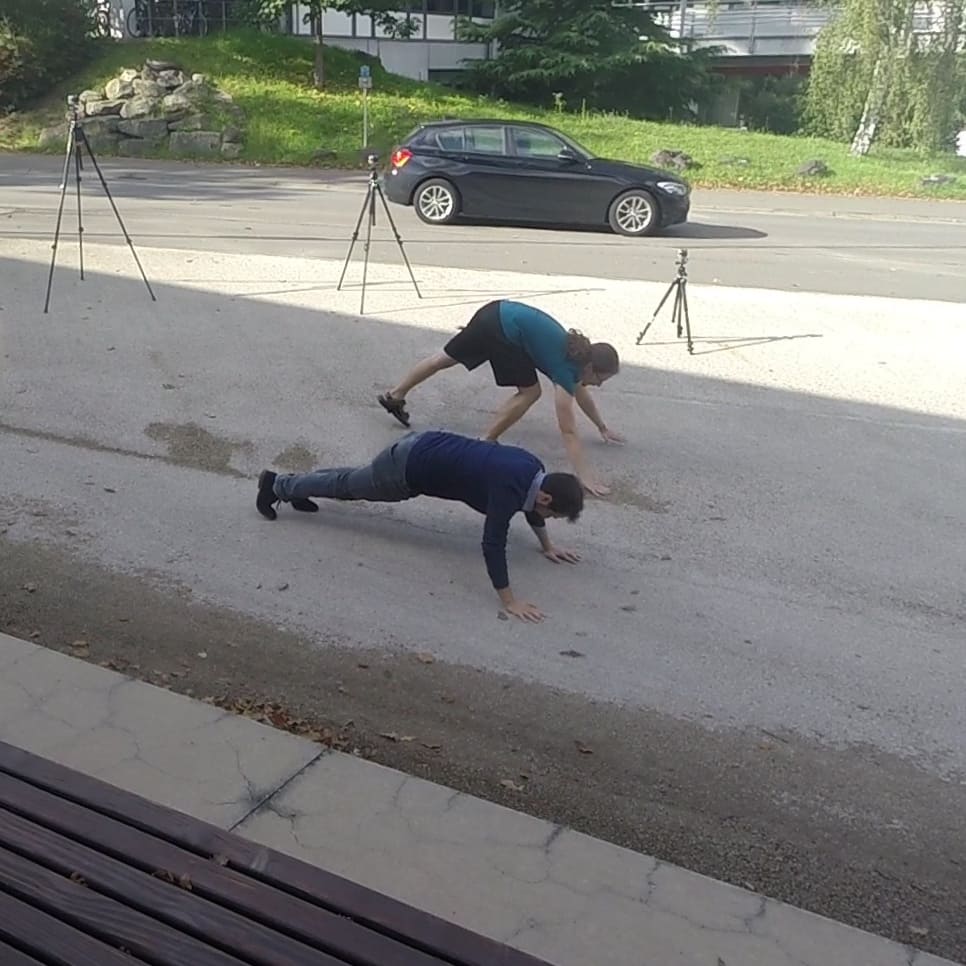}}
\subfloat[SPEC~\cite{kocabas2021spec}]{\includegraphics[width=0.19\textwidth]{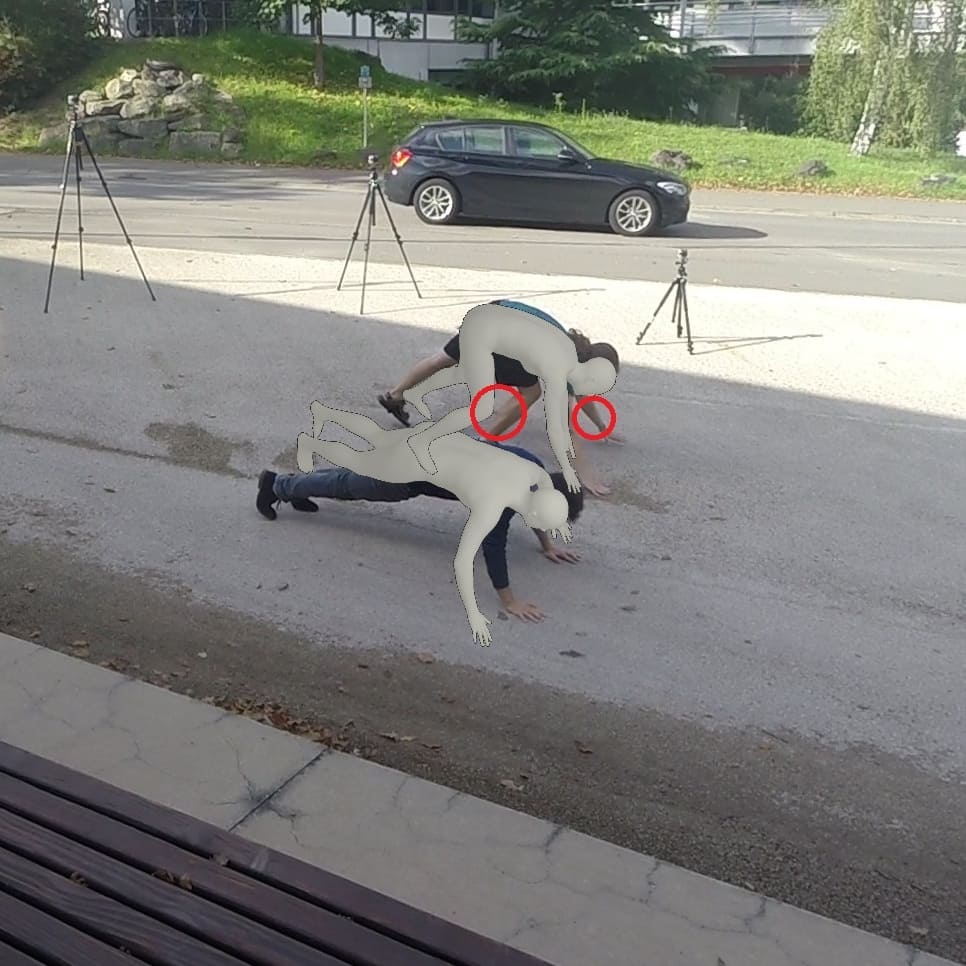}}
\subfloat[ROMP~\cite{sun2021monocular}]{\includegraphics[width=0.19\textwidth]{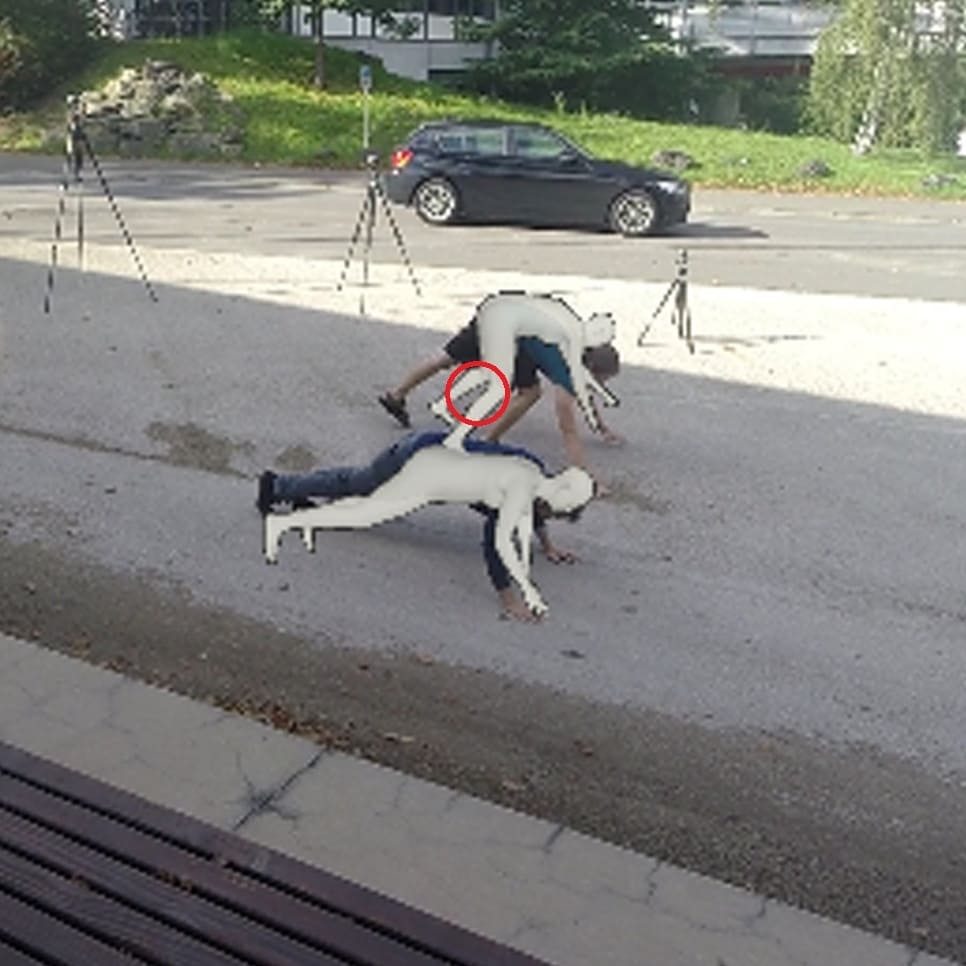}}
\subfloat[PARE~\cite{kocabas2021pare}]{\includegraphics[width=0.19\textwidth]{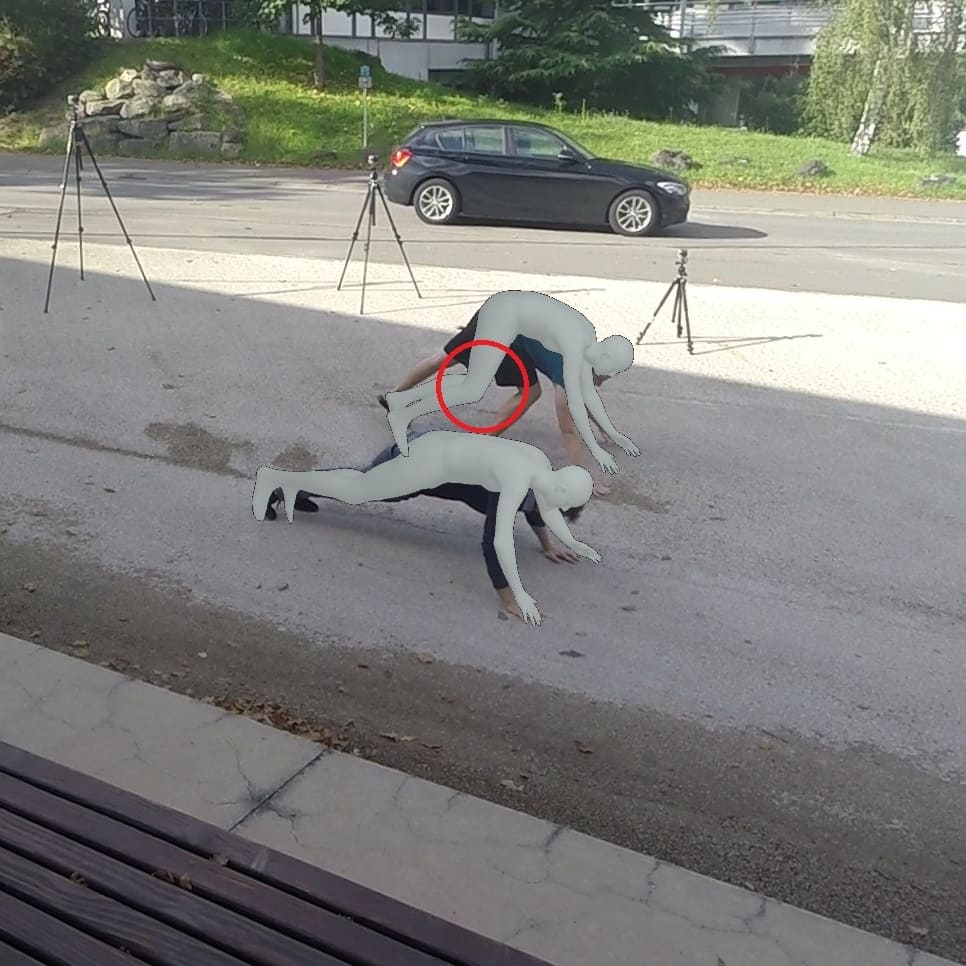}}
\subfloat[\textit{Ours}]{\includegraphics[width=0.19\textwidth]{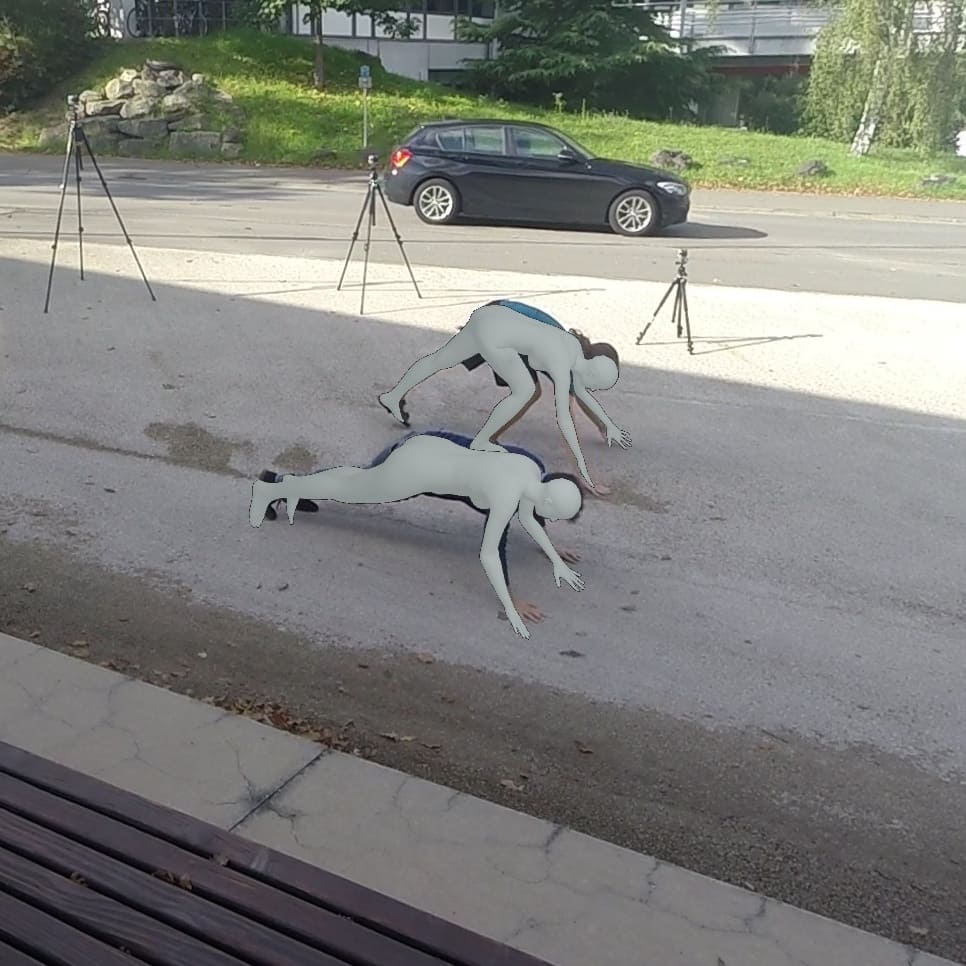}}
\caption{Qualitative comparisons on 3DPW (Row 1), AGORA (Rows 2-3) and MuPoTS (Rows 4-5) datasets. Red circles highlight wrongly estimated parts.}
\label{qualresult}
\end{figure*}

Fig.~\ref{attention} shows the visualization for the attention learned in the relation-aware refinement network $f^\text{Ref}$. It learns the attentions for intra-person parts as in the column 1 and learns the inter-person attentions among at most $N=3$ persons as in columns 2 through 4. From the visualization, we could see that the refinement network $f^\text{Ref}$ refines the initial meshes (in the 6th row) a lot towards refined meshes (in the 7th row). Fig.~\ref{qualresult} shows the qualitative comparisons to competitive state-of-the-arts~\cite{kocabas2021pare,kocabas2021spec,sun2021monocular}. Ours faithfully reconstructs 3D human bodies with diverse artifacts while others frequently fails to capture the details.

\noindent \textbf{Ablation study.} We conduct an ablation study for several design choices. The Table~\ref{tab:ablation study} shows the ablation results on 3DPW dataset: `Ours w.o IK, w/o Ref', `Ours w/o Ref', `Ours w/ positional embedding', `Ours w/o masking input patch' denote our results obtained without inverse kinematics process and refinement module which are 3D skeletons, our results without the refinement module, our results with positional embedding and our results without masking input patches, respectively. From the results, we can see that inverse kinematics and relation-aware refinements consistently increase the accuracy of our pipeline. Furthermore, we decided not to use positional embedding while using the masking input patch scheme. `Ours (N=1)' through `Ours (N=4)' denote experiments conducted by varying the maximum number of persons (i.e. $N$) of the Transformer input. We observe that $N=3$ works best.

\section{Conclusion}
In this paper, we proposed a coarse-to-fine pipeline for the multi-person 3D mesh reconstruction task, which first estimates occlusion-robust 3D skeletons, then reconstructs initial 3D meshes via the inverse kinematic process and finally refines them based on the relation-aware refiner considering intra- and inter-person relationships. By extensive experiments, we find that our idea of delivering the accurate occlusion-robust 3D poses to 3D meshes, and refining initial mesh parameters of interacting persons indeed works: Our pipeline consistently outperforms multiple 3D skeleton-based, 3D mesh-based baselines and each component proposed works meaningfully for the intended scenario. 

\noindent \textbf{Acknowledgements.} This work was supported by IITP grants (No. 2021-0-01778 Development of human image synthesis and discrimination technology below the perceptual threshold; No. 2020-0-01336 Artificial intelligence graduate school program(UNIST); No. 2021-0-02068 Artificial intelligence innovation hub; No. 2022-0-00264 Comprehensive video understanding and generation with knowledge-based deep logic neural network) and the NRF grant (No. 2022R1F1A1074828), all funded by the Korean government (MSIT). 

\clearpage
%
%
\bibliographystyle{abbrv}
\bibliography{ma}
\end{document}